\documentclass[10pt,twocolumn,letterpaper]{article}

\usepackage[pagenumbers]{cvpr} 

\definecolor{cvprblue}{rgb}{0.21,0.49,0.74}
\usepackage[pagebackref,breaklinks,colorlinks,allcolors=cvprblue]{hyperref}

\usepackage{arydshln} 
\usepackage{multirow}
\usepackage{makecell}
\usepackage[most]{tcolorbox}
\usepackage{xcolor}
\usepackage{colortbl}
\usepackage{booktabs} 
\usepackage{bm}
\usepackage{graphicx}
\usepackage[margin=1in]{geometry}
\usepackage{tikz}
\usepackage{amsmath}
\usepackage{nicefrac}
\usepackage{xcolor}
\definecolor{lightgray}{gray}{0.8} 
\usepackage{todonotes}
\usepackage{pdfcomment}

\def\paperID{} 
\def\confName{CVPR}
\def\confYear{2026}

\title{
SVGFusion: A VAE-Diffusion Transformer for Vector Graphic Generation
}

\author{%
  Ximing Xing$^{1}$,
  Juncheng Hu$^{1}$,
  Ziteng Xue$^{1}$,
  Jing Zhang$^{1}$,
  Buyu Li$^{2}$,
  Sheng Wang$^{2}$, \\
  Dong Xu$^{3}$,
  Qian Yu$^{1}$\thanks{Corresponding author} \\
  $^{1}$Beihang University \quad $^{2}$Bambu AI \quad $^{3}$The University of Hong Kong \\
  \texttt{\{ximingxing, qianyu\}@buaa.edu.cn} \quad
  \texttt{dongxu@cs.hku.hk}
}

\begin{document}
\twocolumn[{
\renewcommand\twocolumn[1][]{#1}
\maketitle
\vspace{-10pt} 
\centering     
\captionsetup{type=figure}
\includegraphics[width=1.0\textwidth]{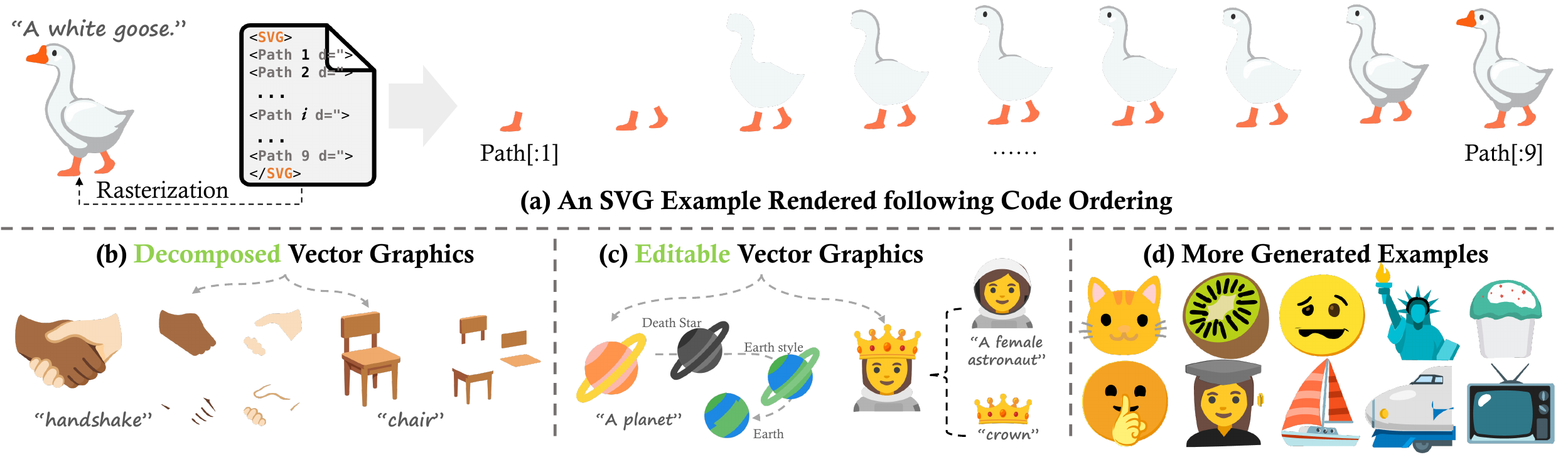}
\captionof{figure}{
\textbf{An overview of our SVGFusion for text-to-SVG generation.}
Our proposed method, SVGFusion can generate SVGs with \textbf{(a)} reasonable construction, \textbf{(b)} a clear and systematic layering structure, and \textbf{(c)} high editability. \textbf{(d)} illustrate more examples generated by our SVGFusion.
}
\label{fig:teaser}
\vspace{15pt} 
}]

\begin{abstract}
Generating high-quality Scalable Vector Graphics (SVGs) from text remains a significant challenge. Existing LLM-based models that generate SVG code as a flat token sequence struggle with poor structural understanding and error accumulation, while optimization-based methods are slow and yield uneditable outputs. To address these limitations, we introduce SVGFusion, a unified framework that adapts the VAE-diffusion architecture to bridge the dual code-visual nature of SVGs. Our model features two core components: a Vector-Pixel Fusion Variational Autoencoder (VP-VAE) that learns a perceptually rich latent space by jointly encoding SVG code and its rendered image, and a Vector Space Diffusion Transformer (VS-DiT) that achieves globally coherent compositions through iterative refinement. Furthermore, this architecture is enhanced by a Rendering Sequence Modeling strategy, which ensures accurate object layering and occlusion. Evaluated on our novel SVGX-Dataset comprising 240k human-designed SVGs, SVGFusion establishes a new state-of-the-art, generating high-quality, editable SVGs that are strictly semantically aligned with the input text.
\end{abstract}
\section{Introduction}
\label{sec:intro}
Scalable Vector Graphics (SVGs) are a cornerstone of modern digital design due to their resolution-independence, which allows them to be scaled to any size without loss of detail. Furthermore, their programmatic structure affords a high degree of editability, enabling designers to precisely modify individual graphic elements. Consequently, SVGs are widely used in applications such as web design, user interfaces, and the creation of icons, logos, and emojis.

The task of Text-to-SVG Generation has garnered increasing attention in recent years. 
An SVG has a dual nature: it is simultaneously a structured, XML-based \textbf{code} and, upon rendering, a \textbf{visual} graphic. 
Existing generation methods can be broadly categorized by which aspect they prioritize.
Optimization-based methods, such as those described in \cite{clipdraw_frans_2022, evolution_tian_2022, Clipasso_vinker_2022, CLIP_radford_2021, CLIPVG_song_2023, vectorfusion_jain_2023, diffsketcher_xing_2023, svgdreamer_xing_2023}, approach the task from a \textbf{visual} perspective. 
These approaches iteratively refine a set of vector parameters by using a differentiable rasterizer \cite{diffvg_Li_2020} to compare the SVG's rendered appearance against guidance from vision-language models like CLIP \cite{CLIP_radford_2021} or Stable Diffusion \cite{ldm_Rombach_2022}. While capable of high visual fidelity, this process is computationally intensive, supports only a limited subset of differentiable SVG commands (e.g., Bézier curves), and produces poorly structured graphics with intertwined primitives that are difficult to edit.

\newcommand{\shadowtoken}[1]{%
  \tcbox[
    on line,              
    arc=2pt,              
    colback=gray!15,      
    boxrule=0pt,
    boxsep=0pt,           
    left=1pt, right=1pt,  
    top=0pt, bottom=0pt,  
    drop fuzzy shadow={gray!50} 
  ]{\texttt{\strut#1}}
}

\begin{figure}
\centering
\includegraphics[width=1\linewidth]{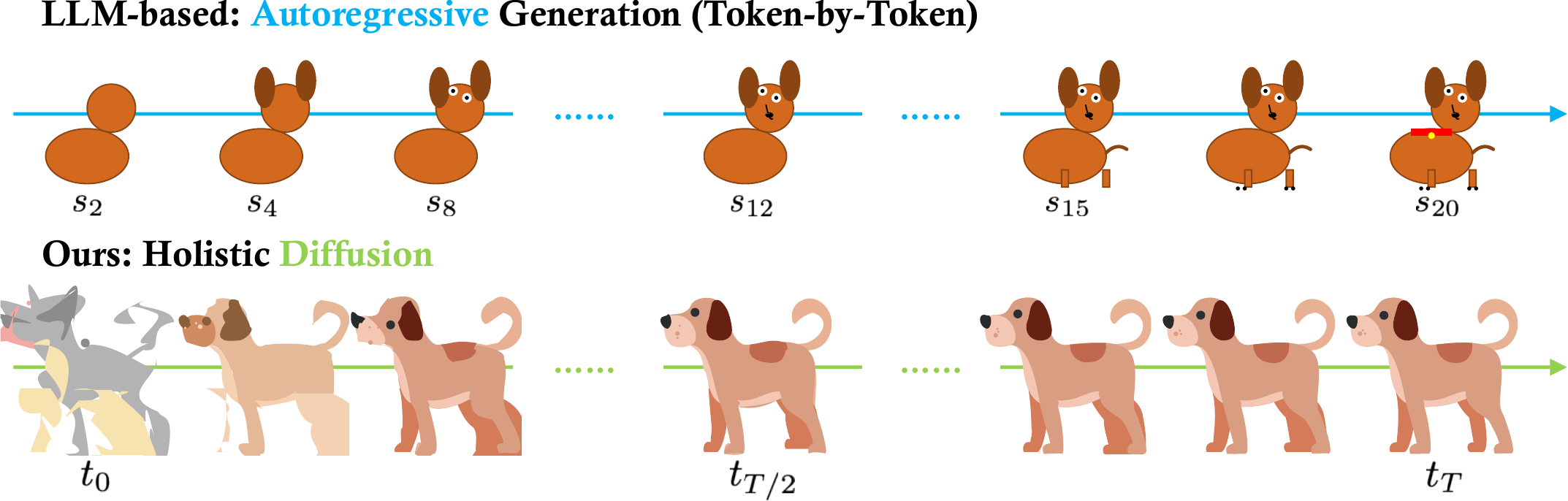}
\vspace{-2em}
\caption{
\textbf{Comparison of autoregressive generation (LLM-based) and diffusion-based generation (Ours).} 
LLM-based models generate vector graphics sequentially, predicting one token at a time based on prior outputs, which can lead to error accumulation and limited global coherence. In contrast, our SVGFusion leverages diffusion-based generation, which operates in a continuous vector space, enabling more holistic and globally consistent SVG synthesis while mitigating the limitations of discrete token-based generation.
}
\vspace{-1.5em}
\label{fig:supp_llm_vs_ddpm}
\end{figure}

In contrast, language-model-based methods~\cite{sketchrnn_david_2018, deepsvg_carlier_2020, deepvecfont_wang_2021, deepvecfontv2_wang_2023, iconshop_wu_2023, strokenuwa_tang_2024}, especially those based on Large Language Models (LLMs)~\cite{llm4svg_xing_2024, omnisvg_yang_2025, internsvg_wang_2025}, have recently become the mainstream approach, treating an SVG as \textbf{code} and framing the task as a sequential generation problem. 
However, we argue that the autoregressive (AR) nature of these LLM-based models introduces fundamental limitations that are ill-suited for SVG generation.
(1) \textbf{Insufficient and unstructured representation}. LLM-based models process SVGs as a flat string of tokens, \textit{e.g.}, \texttt{\shadowtoken{<path}\shadowtoken{,}\shadowtoken{ d}\shadowtoken{="}\shadowtoken{ M}\shadowtoken{,}\shadowtoken{ 10}\shadowtoken{,}\shadowtoken{ 10}\shadowtoken{,}\shadowtoken{ L}\shadowtoken{,} ...">}. This representation is insufficient and loses the inherent structure of the graphic. The model does not intrinsically know that \shadowtoken{10} and \shadowtoken{10} form a coordinate pair, which belongs to a \texttt{moveto} command, which is part of a single path. 
(2) \textbf{Lack of global visual coherence}. LLM-based model lacks a holistic view of the canvas during generation. Its predictions are conditioned only on preceding sequence of code, such process often leads to poor composition and a general lack of spatial harmony, as the model cannot `see' how a new shape will fit into the complete output. 
(3) \textbf{Irreversible accumulated errors}. For LLM-based models, a single mistake early in the sequence, like misplaced coordinate or malformed path command, will become a permanent part of the context for all subsequent predictions, often triggering a cascade of failures that results in corrupted SVGs (See Fig.~\ref{fig:supp_llm_vs_ddpm}).


In this work, we propose \textbf{SVGFusion}, a novel Text-to-SVG generation model that adapts the powerful VAE-Diffusion architecture from the image domain to overcome the limitations of LLM-based approaches. Our model consists of two synergistic components: a Vector-Pixel Fusion Variational Autoencoder (VP-VAE) and a Vector Space Diffusion Transformer (VS-DiT). 
\textbf{\textit{First}}, VP-VAE learns a continuous and structured latent space for SVGs. Instead of processing lengthy, discrete sequences of tokens as in LLM-based models, VP-VAE learns to encode the entire SVG by vectorizing each individual path and primitive as a distinct entry in a structured matrix, which is then holistically mapped into the latent space. Therefore, this approach imbues the model with an intrinsic understanding of the graphic's compositional structure.
\textbf{\textit{Second}}, once the VAE has established a robust latent space, a diffusion model, i.e., DiT~\cite{dit_peebles_2023} generates new latent codes that can be decoded into SVGs. The diffusion process is inherently \textit{global} and \textit{iterative}. Specifically, during the diffusion process, it evaluates the entire latent representation at each step, enabling the model to consider the global composition of the SVG. The self-attention layers within the DiT can effectively capture the global context of the latent representation. Besides, the multi-step denoising process allows for continuous refinement. Inaccuracies or inconsistencies introduced in one step can be corrected in subsequent steps as the model converges on a final output.
Therefore, our SVGFusion can effectively avoid the issues raised in LLM-based models.

To learn a comprehensive latent representation, we introduce \textbf{Vector-Pixel Fusion Encoding (VPFE)} into VP-VAE. 
The VPFE component leverages the dual nature of SVGs by learning a latent space from both their symbolic code and their rendered, pixel-based images. This fusion results in a more meaningful, robust, and visually coherent latent space.
Specifically, the VP-VAE encoder has two branches: one for the structured code and another for deep visual features extracted from the rendered image using a model like DINOv2~\cite{dinov2_oquab_2024}. 
The rendered image provides a powerful perceptual signal, enabling the encoder to recognize that syntactically different SVGs can be visually similar. 
For instance, the choice of primitive type or the sequence of commands can vary significantly while yielding a similar rendered appearance.
A standard VAE, learning solely from SVG code, would incorrectly map these variations to distant points in the latent space. In contrast, by being guided by the shared visual context, our VP-VAE is encouraged to map them to more proximate points in the latent space, better reflecting their perceptual similarity.


Furthermore, to ensure our model understands the sequential logic embedded within SVG code, we introduce a novel  \textbf{Rendering Sequence Modeling}  strategy. This sequential dependency is critical because SVGs adhere to a painter's model, where primitives defined later in the code are rendered on top of, and may occlude, those defined earlier. Altering this order can corrupt the final image or render it incomplete. Our strategy addresses this by training the model on sequences of incrementally constructed SVGs and their corresponding renderings. This process explicitly teaches the model about layering and occlusion, equipping SVGFusion to generate SVGs with coherent structures.


To facilitate this research, we constructed \textbf{SVGX-Dataset}, a new, large-scale dataset of approximately 240,000 high-quality, human-designed SVGs curated from various online sources. We also developed an automated pre-processing pipeline to clean and simplify the SVGs losslessly.
Our contributions are threefold:
\begin{itemize}
\item \textbf{A Novel SVG Generation Framework:} 
We introduce SVGFusion, a new model that successfully adapts the powerful VAE-DiT architecture from the T2I domain to text-to-SVG code generation, producing high-quality vector graphics.
\item \textbf{SVG-Specific Architectural Innovations:} 
We propose VP-VAE that learns a robust latent space by fusing code and pixel information. We also introduce a Rendering Sequence Modeling strategy that enables the model to understand the constructive logic of SVGs, enhancing the quality of the output.
\item \textbf{A Large-Scale Dataset and Benchmark:}  
We construct \textit{SVGX-Dataset},  a comprehensive collection of ~240,000 high-quality SVGs. Using this dataset, we conduct extensive experiments that validate the effectiveness of SVGFusion and establish a new state-of-the-art benchmark for the task.
\end{itemize}

\begin{figure*}[t!]
\centering
\includegraphics[width=1.0\textwidth]{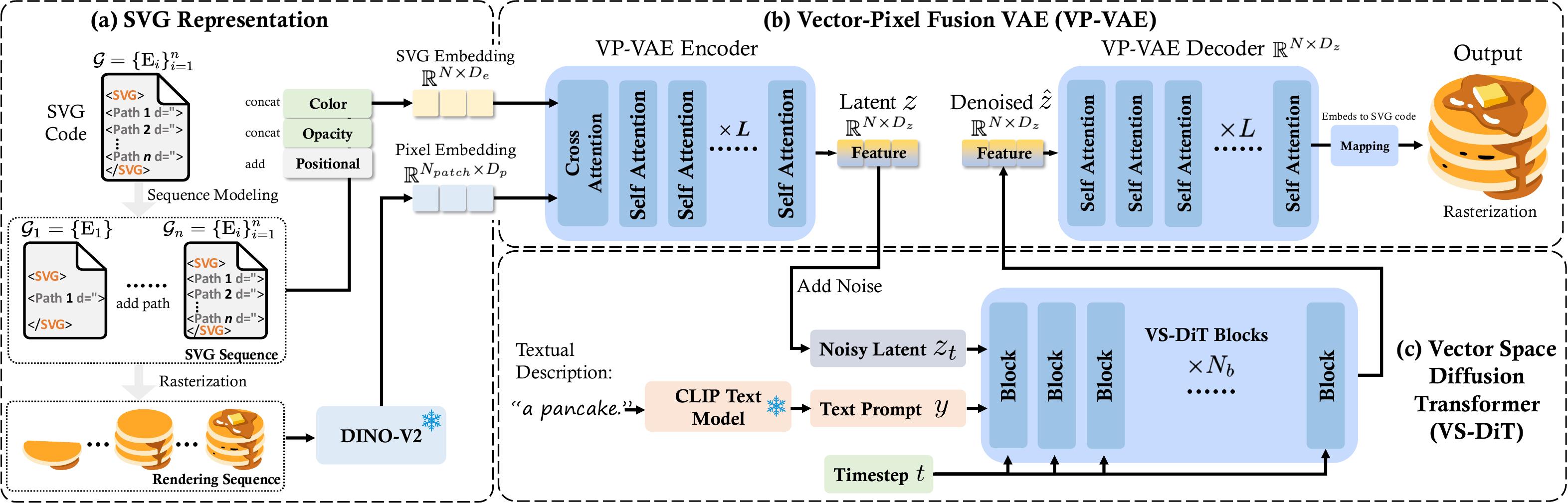}
\vspace{-2em}
\caption{
\textbf{An Overview of SVGFusion.}
(a) The pipeline begins with the representation of SVGs, where XML-defined SVG code is converted into an SVG embedding (Sec.~\ref{sec:svg_rep}).
(b) We first train a Vector-Pixel Fusion Variational Autoencoder (VP-VAE, Sec.~\ref{sec:vp_vae}) with a transformer-based architecture to learn a continuous latent space for SVGs by incorporating features from both SVG codes and their rendered images.
(c) The Vector Space Diffusion Transformer (VS-DiT, Sec.~\ref{sec:dit}) is then trained within the learned latent space to generate new latent codes conditioned on input text descriptions.
} \label{fig:pipeline}
\vspace{-1em}
\end{figure*}
\section{Related Work}
\label{sec:related_work}
\subsection{Vector Graphics Generation}
Scalable Vector Graphics (SVGs) are widely used in design due to their geometric manipulability, resolution independence, and compactness. Early SVG generation methods train neural networks to output predefined SVG commands and attributes~\cite{sketchrnn_david_2018, svgvae_lopes_2019, deepsvg_carlier_2020, im2vec_reddy_2021, iconshop_wu_2023, strokenuwa_tang_2024} using RNNs, VAEs, or Transformers. However, their capability to model complex and diverse vector graphics is limited by the scarcity of large-scale training data.

Compared with raster image generation, which benefits from datasets like ImageNet~\cite{imagenet_deng_2009}, available vector datasets remain narrow in domain—primarily icons~\cite{figr8_clouatre_2019}, emojis~\cite{notoemoji_google_2022}, and fonts~\cite{deepvecfont_wang_2021}. As an alternative to direct SVG generation, optimization-based methods iteratively refine vector parameters to match a target image.

DiffVG~\cite{diffvg_Li_2020} introduced a differentiable rasterizer that enables gradient-based SVG optimization, later extended by works combining differentiable rasterization with VLMs such as CLIP~\cite{CLIP_radford_2021} for text-guided vector synthesis~\cite{clipdraw_frans_2022, Clipasso_vinker_2022, evolution_tian_2022, LIVE_Ma_2022, CLIPVG_song_2023, diffsketcher_xing_2023, supersvg_hu_2024, svgneualpath_zhang_2024, xing2024svgdreamer++}. More recently, diffusion models like DreamFusion~\cite{dreamfusion_poole_2023} have inspired vector extensions—VectorFusion~\cite{vectorfusion_jain_2023}, DiffSketcher~\cite{diffsketcher_xing_2023}, and SVGDreamer~\cite{svgdreamer_xing_2023}—which produce higher-quality sketches and icons but still face challenges in editability, geometry redundancy, and visual consistency. Hybrid methods~\cite{NIVeL_thamizharasan_2024, svgneualpath_zhang_2024} introduce geometric constraints to refine paths but remain confined to SDS-optimized structures. VecFusion~\cite{vecfusion_thamizharasan_2024} advances image-conditioned diffusion for vector fonts, yet its scope is limited to font synthesis.

In contrast, SVGFusion proposes a scalable, continuous vector-space generative framework that moves beyond discrete code models and optimization-heavy pipelines, enabling diverse, editable, and high-quality SVG generation.

\subsection{Diffusion Model}
\noindent Denoising diffusion probabilistic models (DDPM)~\cite{diffusion_models_dickstein_2015,EestGrad_song_2019,ddpm_ho_2020,scorebased_song_2021,ADM_dhariwal_2021,iDDPM_nichol_2021,ddim_song_2021,ldm_Rombach_2022,classifierfree_2022_ho} have demonstrated outstanding performance in generating high-quality images.
The diffusion model architecture combined with the language-image pretrained model~\cite{CLIP_radford_2021} shows obvious advantages in text-to-image (T2I) tasks, including GLIDE~\cite{GLIDE_2022_nichol}, Stable Diffusion~\cite{ldm_Rombach_2022}, DALL·E 2~\cite{DALLE2_2022_ramesh}, Imagen~\cite{imagen_2022_saharia} and DeepFloyd IF~\cite{deepfloydif_stability_2023}, SDXL~\cite{sdxl_podell_2024}.
The progress achieved by T2I diffusion models~\cite{GLIDE_2022_nichol,ldm_Rombach_2022,DALLE2_2022_ramesh,imagen_2022_saharia} also promotes the development of a series of text-guided tasks, such as text-to-3D~\cite{dreamfusion_poole_2023,sjc_wang_2023,prolificdreamer_wang_2023} and text-to-video~\cite{vdm_ho_2022,animatediff_guo_2024,makeavideo_singer_2023,sora_liu_2024}.

Recent efforts such as DreamFusion~\cite{dreamfusion_poole_2023} explores text-to-3D generation by exploiting a Score Distillation Sampling (SDS) loss derived from a 2D text-to-image diffusion model~\cite{imagen_2022_saharia, ldm_Rombach_2022} instead, showing impressive results.
In addition, Sora~\cite{sora_liu_2024} based on the latent diffusion model~\cite{dit_peebles_2023} has made amazing progress in the field of video generation.
Recently, the architecture of diffusion models has been shifting from U-Net~\cite{ADM_dhariwal_2021} architectures to transformer-based architectures~\cite{uvit_bao_2023,dit_peebles_2023,sit_ma_2024}, narrowing the gap between image generation and language understanding tasks. 
In this work, we extend the diffusion transformer to the domain of vector graphics, enabling the synthesis of vector graphics.
We also demonstrate the potential of the proposed method in vector design.
However, the absence of a scalable foundation model for vector graphics has significantly hindered the development of this field for broader applications.
To address this, we propose SVGFusion, a scalable foundation model based on vector space design.
\section{Methods}
\label{sec:method}
Our task is to generate SVGs from input text prompts. As illustrated in Fig.~\ref{fig:pipeline}, our method first trains a Vector-Pixel Fusion Variational Autoencoder (VP-VAE) to learn a latent space $\mathcal{Z}$ for SVGs. Next, a Vector Space Diffusion Transformer (VS-DiT) is trained within this latent space to generate new latent codes conditioned on text prompts.
Once trained, given an input text and a randomly sampled latent code, our model produces an SVG that is semantically aligned with the text.
In this section, we first describe the process of converting SVG code into SVG embeddings (Sec.\ref{sec:svg_rep}), followed by explanation of VP-VAE (Sec.\ref{sec:vp_vae}) and VS-DiT (Sec.~\ref{sec:dit}).


\begin{figure}
\centering
\includegraphics[width=1.0\linewidth]{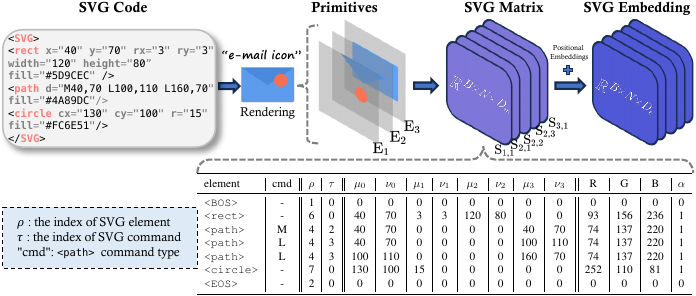}
\vspace{-2em}
\caption{
\textbf{Illustration of the SVG embedding process}. SVG code is initially converted into a matrix representation that includes geometric attributes, colors, and opacity. This matrix is subsequently mapped into a tensor via SVG embeddings.
} \label{fig:code2emb}
\vspace{-1.5em}
\end{figure}
\subsection{SVG Representation}
\label{sec:svg_rep}
Vector graphics consist of machine instructions composed of a series of XML elements (\textit{e.g.} \texttt{<path>} or \texttt{<rect>}), commands (\textit{e.g.} \texttt{M}, \texttt{C} in \texttt{<path>} element), and attributes (\textit{e.g.} \texttt{d}, \texttt{r} or \texttt{fill}). 
Inspired by prior works~\cite{deepsvg_carlier_2020,deepvecfont_wang_2021}, we transform these instructions into a structured, rule-based matrix representation.
We formally define an SVG code as a collection of $n$ vector primitives: SVG $\mathcal{G}=\{\mathrm{E}_1, \mathrm{E}_2, \dots, \mathrm{E}_n\}=\{\mathrm{E}_i\}_{i=1}^{n}$, where each $\mathrm{E}_i$ corresponds to an SVG element such as \texttt{<path>}, \texttt{<rect>}, \texttt{<circle>}, etc.
Each element is defined as: $\mathrm{E}_i=\{\mathrm{S}_{i,j},\mathrm{F}_{i,j},\alpha_{i,j} \}_{j=1}^{M_i}$, where $\mathrm{S}_{i,j}$ is the $j$-th \textit{command} in the $i$-th element, $\mathrm{F}_i \in \{r,g,b\}$ and $\alpha_i \in [0,1]$ correspond to the color property  and the visibility of the $i$-th element, respectively.
$M_i$ indicates the total number of commands in $\mathrm{E}_i$. 
Thus, the total number of primitive commands in an SVG is $N = \sum_{i=1}^n M_i$. 
In our work, we set $N=512$, determining the maximum number of commands in an SVG representation.
Notably, the $\texttt{<path>}$ element consists of multiple commands, while other elements typically contain only a single command.

\noindent\textbf{SVG Embedding.}\quad Figure~\ref{fig:code2emb} illustrates the process of converting an SVG code into an SVG embedding. We begin by transforming each primitive into a vector representation, which we then organize into an SVG matrix.
Specifically, each command $\mathrm{S}_{i,j}$ is represented as $\mathrm{S}_{i,j}=(\rho,\tau,\mu_0,\nu_0,\mu_1,\nu_1,\mu_2,\nu_2,\mu_3,\nu_3)_j^i$, where $\rho$ and $\tau$ represent the type index of an element and command, respectively.
Taking the element \texttt{<rect>} as an example, its element index $\rho$  is 6, and its command index $\tau$ is 0 (as it only has no \texttt{<path>} command), its \( (\mu_i, \nu_j)_{i=0,j=0}^{3,3} \) correspond to \texttt{x}, \texttt{y}, \texttt{rx}, \texttt{ry}, \texttt{width} and \texttt{height}. Further details are provided in Table~\ref{tab:svg_commands} in Supplementary.
After converting into a matrix, we apply an embedding layer to transform $\rho$ and $\tau$ from discrete indices into continuous representations while normalizing coordinates and colors in the matrix.  
Finally, we obtain the SVG embedding by adding the positional embedding. 

Compared to previous approaches~\cite{deepsvg_carlier_2020,deepvecfont_wang_2021,iconshop_wu_2023}, our method supports a broader range of SVG primitives, including more elements (\textit{e.g.}, \texttt{<circle>}, \texttt{<rect>}) and commands (\textit{e.g.} \texttt{Q}, \texttt{A} for \texttt{<path>}).
A full list of primitives supported by our model is shown in Table~\ref{tab:svg_commands} of Supp.~\ref{sec:supp_primitive_type}.
This enhancement significantly improves the model’s ability to learn from real-world data, making the generated SVGs more structured and editable.

\noindent\textbf{Rendering Sequence Modeling.}\quad
To enhance the construction logic of the generated SVGs, we introduce a Rendering Sequence Modeling strategy that enables the model to learn the creation logic of SVG rendering.
During training, we represent the SVG as a progressive sequence of drawing steps. Each step incrementally adds new primitives to the SVG, 
simulating the progressive construction along the batch dimension. 
As illustrated with the cat example in Fig.~\ref{fig:pipeline}, 
each SVG is processed into a batch of data of size $B$, where each sample corresponds to a different stage of creation, ranging from  initial primitives to the complete SVG. 
This allows the model to observe the process of SVG creation within a single training iteration, thereby fostering a deeper understanding of the logical layers involved in SVG.

\begin{figure}
\centering
\includegraphics[width=1.0\linewidth]{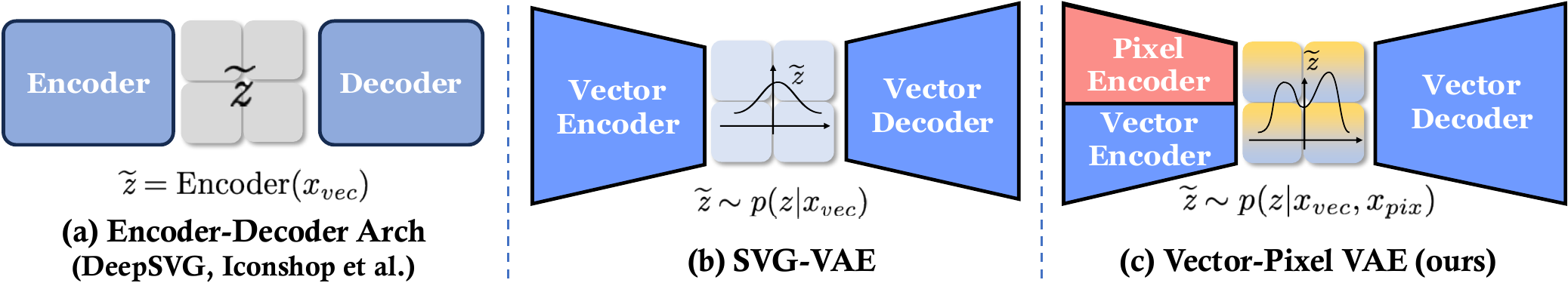}
\vspace{-1.5em}
\caption{
\textbf{Different SVG encoding paradigms.}
DeepSVG performs dimensionality compression, SVG-VAE learns a variational vector latent, and our VP-VAE models a joint vector–pixel latent that fuses visual and code representations.
} \label{fig:rep_compare}
\vspace{-1.5em}
\end{figure}
\subsection{Vector-Pixel Fusion VAE}
\label{sec:vp_vae}
The Vector-Pixel Fusion VAE (VP-VAE) is designed to learn a latent space for vector graphics with a transformer-based VAE architecture, which includes an encoder and a decoder. Different from previous SVG VAE architecture (Fig.~\ref{fig:rep_compare}), VP-VAE is designed to learn from both structural and visual features. Specifically, VP-VAE involves two key innovations in the encoder process: \textit{1)} A Vector–Pixel Fusion Encoder integrates information from the SVG code and its rendered image, enabling the model to jointly learn geometric and visual features. \textit{2)}\ A sequence modeling strategy allows the model to understand how vector graphics are progressively constructed, ensuring the resultant SVGs with more reasonable construction.

\begin{figure}
\centering
\includegraphics[width=1.0\linewidth]{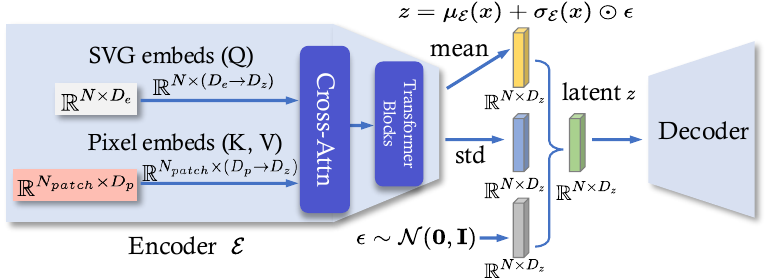}
\vspace{-2em}
\caption{
\textbf{Illustration of the Vector-Pixel Fusion Encoding}. The VP-VAE encoder integrates the SVG embeddings ($Q$) with pixel embeddings ($K$, $V$) using a cross-attention layer. After processing through $L$ self-attention layers, the encoded features are mapped to a latent space, where the mean and standard deviation are computed for a probabilistic representation. 
A latent variable $\bm{z}$ is sampled using the reparameterization trick and then passed to the decoder for further processing.
For clarity, the batch dimension $B$ has been omitted.
} \label{fig:vpvae}
\vspace{-1.5em}
\end{figure}

\noindent\textbf{Vector-Pixel Fusion Encoding.}\quad
As shown in Fig.~\ref{fig:vpvae}, the VP-VAE encoder $\mathcal{E}(\cdot)$ takes both SVG embeddings $\bm{E}_{\mathrm{emb}} \in \mathbb{R}^{N \times D_e}$
and Pixel embeddings $\bm{E}_{\mathrm{pix}} \in \mathbb{R}^{N_{patch} \times D_p}$. 
The pixel embeddings are obtained by using a pretrained DINOv2~\cite{dinov2_oquab_2024} to extract high-level visual features from the rendered images.
To align these embeddings in the latent space, we first project them onto a common dimension $D_z$ using separate linear layers:
$\bm{E}_\mathrm{emb}^{\prime} = \bm{W}_{\mathrm{proj}}^e \bm{E}_\mathrm{emb}^{\top}$, $\bm{E}_\mathrm{pix}^{\prime} = \bm{W}_{\mathrm{proj}}^p \bm{E}_\mathrm{pix}^{\top}$, where $\bm{W}_{\mathrm{proj}}^e \in \mathbb{R}^{D_z \times D_e}, \bm{W}_{\mathrm{proj}}^p \in \mathbb{R}^{D_z \times D_p}$.
The transformed embeddings are then subjected to a cross-attention layer, where the SVG embeddings act as queries ($Q$), while the pixel embeddings act as keys ($K$) and values ($V$). 
This enables the model to effectively integrate geometric and visual features, enhancing its representation of complex vector structures. Finally, the fused representation is processed through the subsequent self-attention layers and mapped to a latent distribution, from which the latent variable $\bm{z}$ is sampled.

\noindent\textbf{The Architecture of VP-VAE.}\quad
The encoder of the VP-VAE consists of one cross-attention layer followed by $L$ self-attention layers. 
The final output, latent code $\bm{z}$, from the encoder is designed to encapsulate both visual and geometric features of the SVG. 

The decoder of the VP-VAE mirrors the structure of the encoder but omits the cross-attention layer. 
Given a latent code $\bm{z}$, the decoder reconstructs its corresponding SVG embeddings: $\bm{E}_{\mathrm{rec}}=\mathcal{D}(\bm{z})$, where $\mathcal{D}(\cdot)$ represents the decoder network.
To obtain the final SVG representation, we apply two separate decoding processes: 
\textit{1) Coordinate Mapping:} The reconstructed embeddings containing normalized coordinates $C_{\text{rec}}$ (ranging from $[-1, 1]$) are mapped back to the canvas coordinate system: $C_{\text{coord}}=\nicefrac{(C_{\text{rec}} + 1)}{2} \times \mathrm{V}$, where $\mathrm{V}$ indicates the predefined canvas size.
\textit{2) Element \& Command Mapping:} The SVG elements and commands are recovered from their respective embeddings using a learned embedding-to-token mapping. Specifically, the reconstructed embeddings are passed through an embedding layer, which predicts the discrete SVG elements and commands, effectively reconstructing the full SVG structure.
These steps ensure that the decoded SVG faithfully preserves both the structural and geometric information encoded in the latent space.

\noindent\textbf{VP-VAE Objective.}\quad
To ensure accurate reconstruction of vector primitives, we measure the discrepancy between the predicted primitive $\mathrm{E}_i$ and the ground-truth primitive $\hat{\mathrm{E}}_i$ using the mean squared error (MSE) loss. This loss encourages the model to generate primitives that closely match the original input.
Besides, we impose regularization on the latent space using Kullback-Leibler (KL) divergence, which constrains the learned latent distribution to approximate a standard Gaussian prior. This prevents overfitting and ensures smooth, continuous latent representations, which are crucial for generating diverse and coherent vector structures.
The final  loss is formulated as:
\begin{equation}
\mathcal{L}_{\mathrm{VAE}} = \mathcal{L}_{\mathrm{MSE}}(\mathrm{E}_i, \hat{\mathrm{E}}_i) + \lambda_{\mathrm{KL}}\mathcal{L}_{\mathrm{KL}}
\label{eq:vae}
\end{equation}
\noindent where $\mathcal{L}_{\mathrm{KL}}$ measures the divergence between the learned latent distribution and the prior Gaussian distribution $\mathcal{N}(\bm{0}, \bm{I})$.
The KL term encourages the latent space to remain compact and structured, facilitating smooth interpolation between vector primitives.

\subsection{Vector Space Diffusion Transformer}
\label{sec:dit}
The VP-VAE effectively learns a latent space tailored for vector graphics representation.
Building on this foundation, SVGFusion leverages DiT~\cite{dit_peebles_2023} as its core architecture and performs the diffusion process directly in the vector latent space.
To facilitate interaction between textual features and vector representations, we introduce a multi-head cross-attention layer into the VS-DiT block, inspired by~\cite{pixartalpha_chen_2024}. 
A detailed description of the VS-DiT architecture is provided in Supplementary Sec.~\ref{sec:supp_vs-dit}.
During training, for a given input SVG, we first derive its latent representation using VP-VAE: $\bm{z}=\mathcal{E}(\mathcal{G})$.
The diffusion process then takes place within this latent space, where the noisy latent variable $\bm{z}_t$ is generated as follows: $\bm{z}_t=\alpha_t \bm{z}+\sigma_t\epsilon, \epsilon\sim\mathcal{N}(\bm{0}, \bm{I})$ where $\alpha_t$ and $\sigma_t$ define the noise schedule, parameterized by the diffusion time $t$.
Following Latent Diffusion Models (LDM)~\cite{ldm_Rombach_2022}, our VS-DiT model predicts the noise $\epsilon$ in the noisy latent representation $\bm{z}_t$, conditioned on a text prompt $y$. The training objective is formulated as:
\begin{equation}
\mathcal{L}_{\mathrm{ldm}}=\mathbb{E}_{ \mathcal{E}(\bm{z}), y, t, \epsilon \sim \mathcal{N}(\bm{0}, \bm{I})}
\left[
\left\|\epsilon_{\phi}(\bm{z}_t,t,y) - \epsilon\right\|_{2}^{2}
\right]
\label{eq:ldm}
\end{equation}
\noindent where $t \sim \mathcal{U}(0,1)$.
We randomly set $y$ to zero with a probability of 10\% to apply classifier-free guidance~\cite{classifierfree_2022_ho} during training, which enhances the quality of conditional generation during inference.

\section{Experiments}
\label{sec:experiments}

\begin{figure*}[t]
\centering
\includegraphics[width=1.0\linewidth]{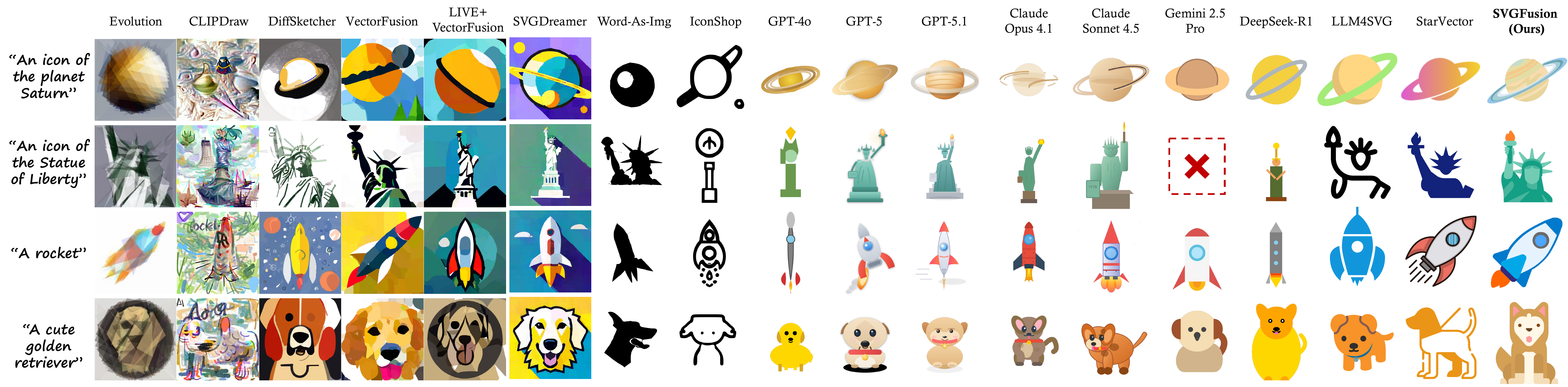}
\vspace{-2em}
\caption{
\textbf{Qualitative Comparison of SVGFusion and Existing Text-to-SVG Methods}. The target SVGs are in the \textit{emoji} style. We use prompt modifiers for the optimization-based approach to encourage the appropriate style: ``minimal flat 2D vector icon, emoji icon, lineal color, on a white background, trending on ArtStation.'' Note that although the visual quality of results generated by optimization-based methods is high, these methods face challenges in decomposing the SVGs for further editing. 
} \label{fig:compare_t2v}
\vspace{-1em}
\end{figure*}

\subsection{SVGX-Dataset}
\label{subsec:svg_data_collect_and_clean}
We introduce \textit{SVGX-Dataset}, a 240K-scale collection of high-quality emoji/icon-style SVGs from Twemoji-Color-Font~\cite{twitter_emoji}, Noto-Emoji~\cite{notoemoji_google_2022}, FluentUI-Emoji~\cite{fluent_ms}, SVG-Repo~\cite{svgrepo}, and Reshot~\cite{reshot_data}. The corpus covers diverse complexities with B\'ezier paths (\texttt{<path>}) and basic primitives (\texttt{<circle>}, \texttt{<rect>}, etc.), spanning black-and-white and color designs across people, animals, objects, and symbols (examples in Fig.~\ref{fig:dataset_examples}, Supplementary).

Web-sourced SVGs often include noise: (1) temporary editor artifacts, (2) suboptimal structure, and (3) unused/invisible elements. We apply a cleaning pipeline that removes redundancies, refines coordinate precision, and standardizes canvases to $224\times224$, reducing file size while preserving visual fidelity (see Fig.~\ref{fig:supp_svg_clean}, Supplementary).

We further analyze names/descriptions via word clouds and compare data pre/post cleaning (Fig.~\ref{fig:dataset_analysis}), confirming a well-structured dataset suitable for training. Additional collection, preprocessing, and analysis details appear in Supplementary Sec.~\ref{sec:supp_svg_data}.

\begin{table}[t]
\centering
\caption{
\textbf{Quantitative Comparison of SVGFusion vs. State-of-the-Art Text-to-SVG Methods}. Top: Optimization-based methods; Middle: language-model-based methods; Bottom: Three model variants of our proposed SVGFusion. Our method uses dpm-solver~\cite{dpmsolver_lu_2022} for 20-step denoising. 
}
\vspace{-0.5em}
\resizebox{1.0\linewidth}{!}{
\begin{tabular}{l|cccc|c}
\toprule
Method / Metric & FID$\downarrow$ & CLIPScore$\uparrow$ & Aesthetic$\uparrow$ & HPS$\uparrow$ & TimeCost$\downarrow$\\
\midrule
Evolution~\cite{evolution_tian_2022} 
& 121.43 & 0.193 & 2.124 & 0.115 & 47min23s \\
CLIPDraw~\cite{clipdraw_frans_2022} 
& 116.65 & 0.249 & 3.980 & 0.135 & 5min10s\\
$\text{DiffSketcher}$\cite{diffsketcher_xing_2023} 
& 72.30 & 0.310 & 5.156 & 0.242 & 10min22s\\
LIVE+VF~\cite{vectorfusion_jain_2023} 
& 82.22 & 0.310 & 4.517 & 0.253 & 30min01s\\
VectorFusion~\cite{vectorfusion_jain_2023} 
& 84.53 & 0.309 & 4.985 & 0.264 & 10min12s\\
Word-As-Img~\cite{wordasimg_Iluz_2023} 
& 101.22 & 0.302 & 3.276 & 0.151 & 5min25s \\
SVGDreamer~\cite{svgdreamer_xing_2023} 
& 70.10 & 0.360 & 5.543 & 0.269 & 35min12s \\
\midrule
SVG-VAE~\cite{svgvae_lopes_2019} 
& 76.22 & 0.190 & 2.773 & 0.101 & 1min\\
DeepSVG~\cite{deepsvg_carlier_2020} 
& 69.22 & 0.212 & 3.019 & 0.114 & 2min\\
Iconshop~\cite{iconshop_wu_2023} 
& 52.22 & 0.251 & 3.474 & 0.140 & 1min03s\\
StrokeNUWA~\cite{strokenuwa_tang_2024} 
& 89.10 & 0.300 & 2.543 & 0.169 & 19s\\
\midrule
\textbf{SVGFusion-S} & 9.62 & 0.373 & 5.250 & 0.275 & 24s\\
\textbf{SVGFusion-B} & 5.77 & 0.389 & 5.373 & 0.281 & 28s\\
\textbf{SVGFusion-L} & \textbf{4.64} & \textbf{0.399} & \textbf{5.673} & \textbf{0.290} & 36s\\
\bottomrule
\end{tabular}
}
\vspace{-0.5em}
\label{tab:compare_t2v}
\vspace{-1.5em}
\end{table}
\subsection{Quantitative Evaluation}
\label{subsec:quantitative}

We compare our proposed method with baseline methods using five quantitative indicators across three dimensions: (1) Visual quality of the generated SVGs, assessed by FID (Fr\'echet Inception Distance)~\cite{FID_Heusel_2017}; (2) Alignment with the input text prompt, assessed by CLIP score~\cite{CLIP_radford_2021}, and (3) Aesthetic appeal of the generated SVGs, measured by Aesthetic score~\cite{aesthetic_christoph_2022} and HPS (Human Preference Score)~\cite{HPS_Wu_2023}. 
To ensure a fair comparison, we also recorded the time cost of different methods to evaluate their computational efficiency.

Comparison results are presented in Table~\ref{tab:compare_t2v}. The methods are categorized into two groups: optimization-based methods (top section of Table~\ref{tab:compare_t2v}) and language-model-based methods (middle section of Table~\ref{tab:compare_t2v}). It is evident that our SVGFusion method surpasses other text-to-SVG methods across all evaluation metrics. This demonstrates the superiority of SVGFusion in generating vector graphics that are more closely aligned with text prompts and human preferences. Notably, compared to optimization-based methods, SVGFusion significantly reduces the time cost, enhancing its practicality and user-friendliness.

\subsection{Qualitative Evaluation}
\label{subsec:qualitative}
Figure~\ref{fig:compare_t2v} presents a qualitative comparison between SVGFusion and existing text-to-SVG methods. The results are aligned with the quantitative results discussed in the previous section.
Specifically, the optimization-based methods, including Evolution~\cite{evolution_tian_2022}, CLIPDraw~\cite{clipdraw_frans_2022}, DiffSketcher~\cite{diffsketcher_xing_2023}, VectorFusion~\cite{vectorfusion_jain_2023}, LIVE~\cite{LIVE_Ma_2022}+VectorFusion~\cite{vectorfusion_jain_2023}, SVGDreamer~\cite{svgdreamer_xing_2023}, and Word-as-Img~\cite{wordasimg_Iluz_2023}, use a differentiable renderer~\cite{diffvg_Li_2020} to backpropagate gradients to vector parameters. 
Evolution~\cite{evolution_tian_2022} and CLIPDraw~\cite{clipdraw_frans_2022} utilize CLIP~\cite{CLIP_radford_2021} as the image prior, while DiffSketcher~\cite{diffsketcher_xing_2023}, VectorFusion~\cite{vectorfusion_jain_2023}, SVGDreamer~\cite{svgdreamer_xing_2023}, and Word-as-Img~\cite{wordasimg_Iluz_2023} adopt T2I diffusion as the image prior. 
Despite their visual advantages, optimization-based methods often produce intertwined vector primitives, diminishing SVG editability. 
Language-model-based methods, such as Iconshop~\cite{iconshop_wu_2023}, GPT4o-latest~\cite{GPT4}, GPT-5~\cite{gpt5}, GPT 5.1~\cite{gpt5.1}, Claude-Opus-4.1~\cite{claude_opus_4.1}, Claude-Sonnet-4.5~\cite{claude_sonnet_4.5}, DeepSeek-R1~\cite{deepseek_r1_2025}, LLM4SVG~\cite{llm4svg_xing_2024} and StarVector~\cite{starvector_rodriguez_2023} rely on language models, and can generate decoupled vector primitives but overly simplistic content.

It is worth noting that although optimization-based methods may produce more realistic or artistic visual effects, they rely on an LDM~\cite{ldm_Rombach_2022} sample as the target for optimization and require a differentiable rasterizer as the medium for this process. Additionally, they depend on differentiable vector primitives as the underlying representation for SVGs. As a result, these methods can only use \texttt{<path>} primitives described by B\'ezier curves. This leads to the need for a large amount of staggered overlapping primitives to closely fit the LDM sample, even for relatively simple shapes. Consequently, even simple regular shapes such as rectangles cannot be described using the corresponding basic shape primitives, thus losing the advantage of SVG's editability, making it difficult to use in real-world scenarios.

\begin{figure}
\centering
\includegraphics[width=1.0\linewidth]{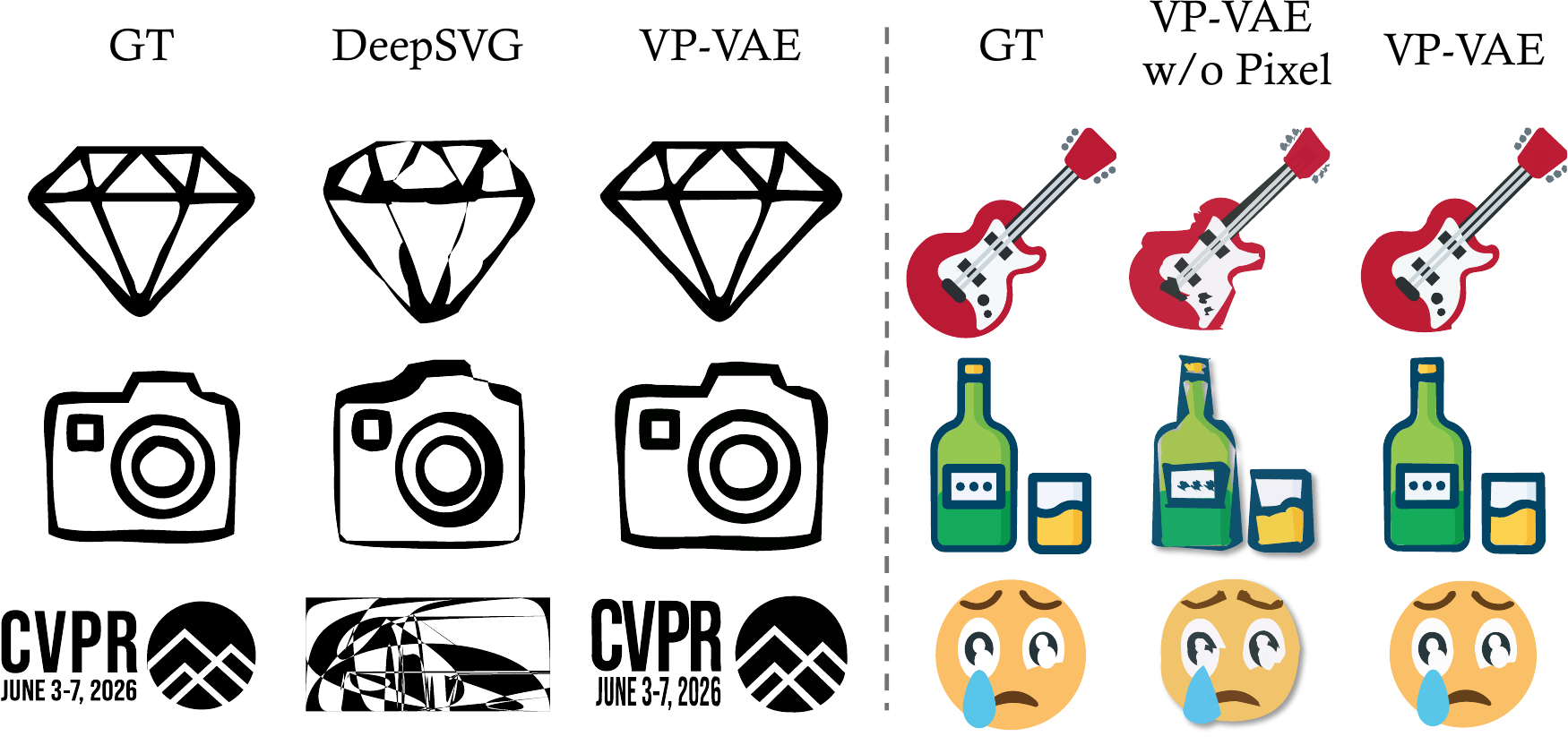}
\vspace{-1.5em}
\caption{
\textbf{SVG reconstruction qualitative comparison.} VP-VAE surpasses DeepSVG~\cite{deepsvg_carlier_2020} on black-and-white SVGs and outperforms the VP-VAE (w/o Pixel) on colored SVGs.
} \label{fig:compare_vae}
\vspace{-1em}
\end{figure}
\begin{figure}[t]
\centering
\includegraphics[width=1\linewidth]{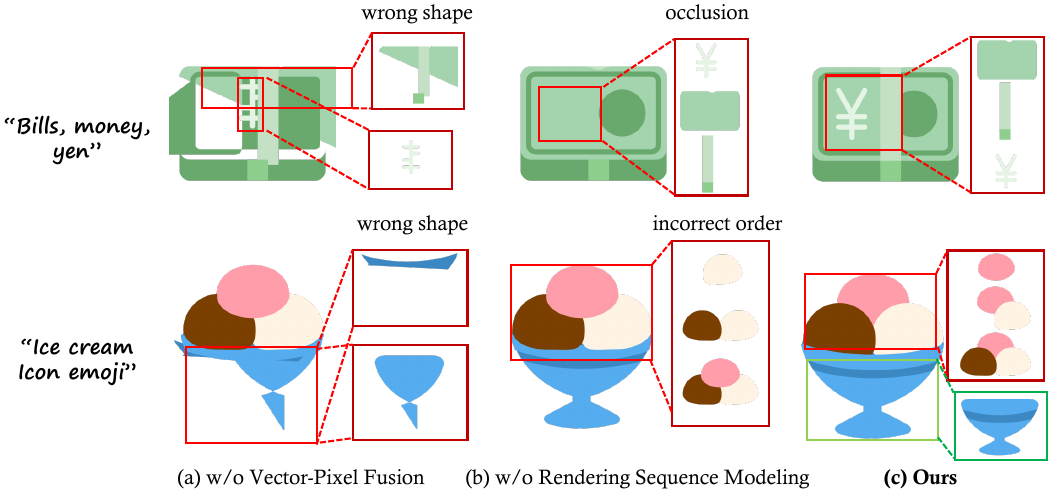}
\vspace{-2em}
\caption{
\textbf{Effects of VP-VAE and Rendering Sequence Modeling.}
\textbf{(a) vs. (c)} demonstrates that VP-VAE cannot accurately reconstruct or generate shapes without the Vector-Pixel Fusion (incorporating DINOv2 features~\cite{dinov2_oquab_2024}).
\textbf{(b) vs. (c)} indicates that employing Rendering Sequence Modeling results in more reasonable SVG outcomes.
} \label{fig:abl_seq}
\vspace{-1.2em}
\end{figure}
\noindent\textbf{Comparison with Large Language Model.}
As illustrated in Fig.~\ref{fig:compare_t2v}, we also compare our proposed SVGFusion with existing state-of-the-art approaches that directly generate SVGs using LLMs. The results indicate that the performance of GPT4o-latest~\cite{GPT4}, GPT-5~\cite{gpt5}, GPT 5.1~\cite{gpt5.1}, Claude-Opus-4.1~\cite{claude_opus_4.1}, Claude-Sonnet-4.5~\cite{claude_sonnet_4.5}, and DeepSeek-R1~\cite{deepseek_r1_2025} in SVG generation is not particularly outstanding. In most cases, they can only use simple shapes to roughly assemble objects, but the positioning of each element lacks harmony. As a result, the overall shapes are overly simplistic, and the visual effects are less satisfactory. Regarding color design, these LLMs struggle to apply colors accurately to each part of the SVG, leading to color schemes that are often neither harmonious nor reasonable. In terms of semantic expression, the SVG code generated by LLMs is too simple to fully capture the meaning conveyed by the input text description. In contrast, our proposed SVGFusion method produces more balanced and harmonious results in terms of shape selection, color matching, and semantic representation. 
In Supplementary Sec.~\ref{sec:supp_compare_llm}, we provide more comparisons of our method with language model-based methods.


\subsection{Ablation \& Analysis}
\label{subsec:ablation}

\begin{table}[t]
\centering
\caption{
\textbf{Comparison of VAE-based SVG representation models.}
We evaluate DeepSVG~\cite{deepsvg_carlier_2020}, vanilla VAE, and our proposed VP-VAE on the FIGR8~\cite{figr8_clouatre_2019} and SVGX datasets, using rFID (↓), SSIM (↑), and PSNR (↑) as metrics.
}
\vspace{-0.5em}
\resizebox{1.0\linewidth}{!}{
\begin{tabular}{lcccccc}
\toprule
\multirow{2}{*}{Model} & \multicolumn{3}{c}{FIGR-8 Dataset} & \multicolumn{3}{c}{SVGX Dataset} \\
\cmidrule(lr){2-4} \cmidrule(lr){5-7}
 & rFID↓ & SSIM↑ & PSNR↑ & rFID↓ & SSIM↑ & PSNR↑ \\
\midrule
DeepSVG & 52.8 & 0.648 & 9.0 & - & - & - \\
VP-VAE w/o Pixel Rep. & 6.7 & 0.852 & 14.8 & 3.1 & 0.876 & 17.5 \\
VP-VAE & \textbf{1.5} & \textbf{0.921} & \textbf{20.7} & \textbf{0.99} & \textbf{0.913} & \textbf{22.8} \\
\bottomrule
\end{tabular}
}
\label{tab:vae_comparison}
\end{table}
\noindent\textbf{Vector–Pixel Fusion Encoding.}\quad
To learn a perceptually meaningful latent space for vector graphics, SVGFusion introduces the \emph{Vector–Pixel Fusion VAE} (VP-VAE), which jointly encodes SVG code structure and its rendered appearance. As shown in Tab.~\ref{tab:vae_comparison}, VP-VAE achieves higher SSIM and PSNR and lower rFID than DeepSVG~\cite{deepsvg_carlier_2020}, demonstrating improved visual fidelity.

Figure.~\ref{fig:compare_vae} provides a qualitative comparison across three paradigms: (i) DeepSVG~\cite{deepsvg_carlier_2020} suffers from jagged contours and missing geometry, (ii) VP-VAE without pixel features captures global shapes but shows color bleeding and texture inconsistencies, while (iii) our full VP-VAE recovers crisp boundaries, coherent colors, and fine-grained details. These results confirm that pixel-level cues are essential for learning a geometry-aware yet perceptually aligned latent space.

Since the diffusion model operates fully within this latent space, reconstruction quality directly impacts SVG generation. We demonstrate this in Fig.~\ref{fig:abl_seq}, where the variant equipped with Vector–Pixel Fusion Encoding produces more faithful shapes and cleaner topology than its counterpart without pixel features, demonstrating that VPFE strengthens both representation learning and generation quality.

\noindent\textbf{Rendering Sequence Modeling.}\quad The Rendering Sequence Modeling strategy in SVGFusion is designed to enhance the construction logic of generated SVGs, making them more editable. In Fig.~\ref{fig:abl_seq}, we evaluate the impact of employing this strategy compared to scenarios where it is omitted. The results clearly show that SVGs generated with Rendering Sequence Modeling exhibit higher visual quality and improved structural integrity.

In the first example, issues arise when shapes with lighter colors are occluded by shapes with darker colors, leading to a poorly represented money icon. In the second example, the creation order of the scoops is critical as their relational positioning greatly influences the visual coherence. 
These examples underscore the effectiveness of our Rendering Sequence Modeling strategy.

\begin{figure}
\centering
\includegraphics[width=0.9\linewidth]{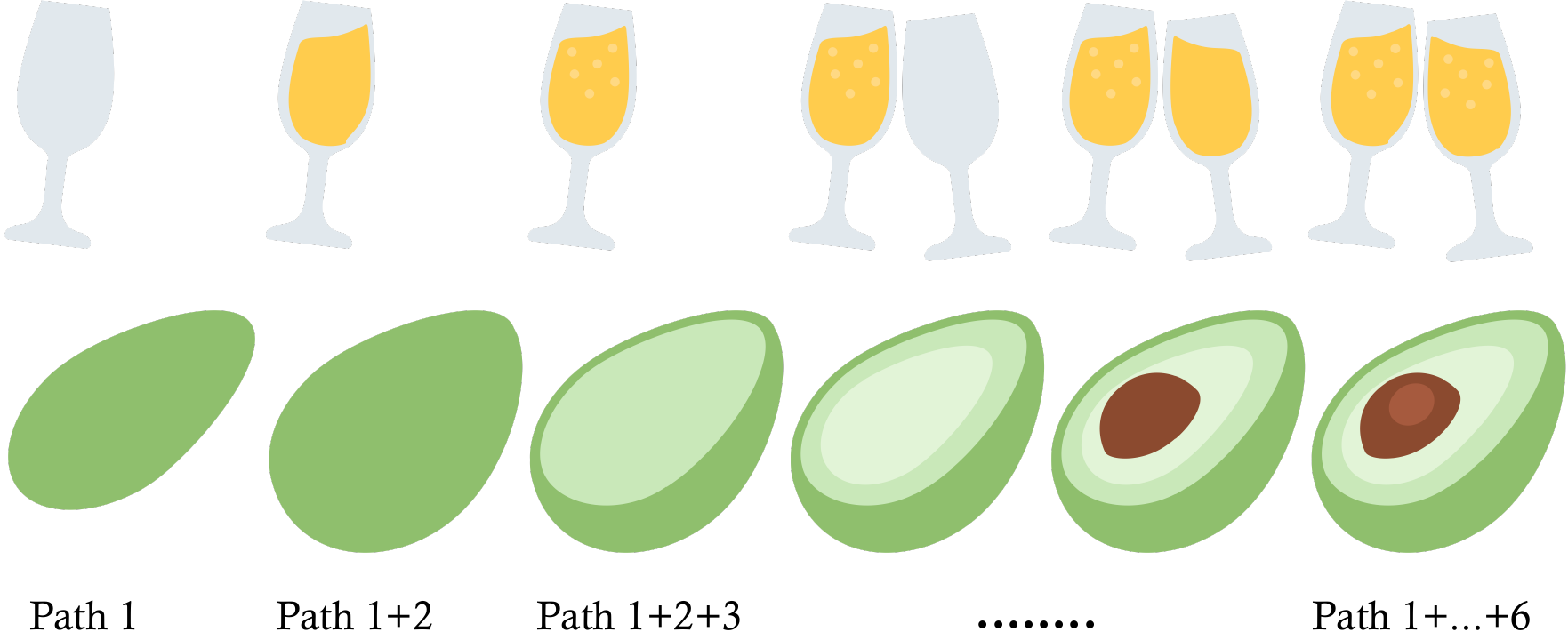}
\vspace{-0.5em}
\caption{
\textbf{Path Rendering Sequence}.
From left to right, each example shows the renderings of accumulated code in each SVG.
} \label{fig:path_render_seq}
\vspace{-1.5em}
\end{figure}
As illustrated in  Fig.~\ref{fig:path_render_seq}, our SVGFusion can generate SVGs using only the necessary primitives, such as \texttt{<circle>} and \texttt{<rect>}, ensuring a compact and structured representation. It allows for flexibility in the edit process: the designer can either start by sketching the general shape and then add local details, or begin with a specific part and gradually add elements to complete the SVG.

\begin{figure}
\centering
\includegraphics[width=\linewidth]{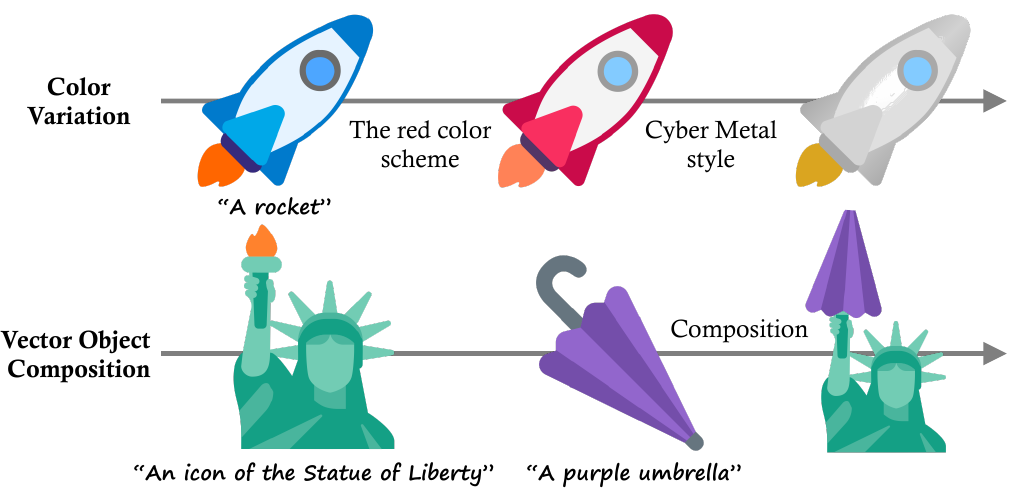}
\vspace{-0.8em}
\caption{
\textbf{The Editability of SVGFusion Results.} 
The SVGs generated by SVGFusion exhibit a clear hierarchical structure, thereby facilitating straightforward edits, such as color modifications (top example) or recomposition into new graphics (bottom example).
} \label{fig:editable}
\vspace{-0.8em}
\end{figure}
\noindent\textbf{Editbility of SVGFusion Results.}
Figure~\ref{fig:editable} shows the editability of the SVG generated by our SVGFusion. 
Since the SVGs we generate have a clean and concise structure, we can easily edit the properties of the primitives, such as their color attributes. For instance, the rocket we generated can be changed from blue to red, or even transformed into a cyber-metal style, simply by adjusting the color attributes.
Furthermore, the capability of edibility empowers users to efficiently reuse synthesized vector elements and create new vector compositions. 
As illustrated in the second example of Fig.~\ref{fig:editable}, our method composes a new vector graphic by replacing the torch with an umbrella.

\section{Conclusion}
\label{sec:conclusion}
In this work, we introduced SVGFusion, a new Text-to-SVG framework that scales to real-world vector graphics without relying on discrete code models or optimization-heavy pipelines. Through the proposed VP-VAE, SVGFusion learns a continuous, visually grounded latent space by fusing SVG structure with DINO-based visual representations, enabling the model to capture both geometric semantics and fine-grained appearance cues. Built upon this latent space, VS-DiT further generates structured SVG elements that reflect realistic creation logic and coherent visual layout. To enable large-scale training, we curated SVGX-Dataset with diverse vector categories and rich annotations. Extensive experiments across reconstruction, generation, and reasoning benchmarks show that SVGFusion achieves superior fidelity, structural consistency, and controllability compared with prior methods.


\clearpage
\renewcommand{\thefigure}{S\arabic{figure}}
\setcounter{figure}{0}
\renewcommand{\thetable}{S\arabic{table}}
\setcounter{table}{0}
\maketitlesupplementary

\appendix

\section*{Overview}
\label{sec:overview}
This supplementary material provides additional implementation details, in-depth analyses, and qualitative results for \textbf{SVGFusion}, organized as follows:

\begin{itemize}
    \item \cref{sec:supp_implement_detail}: \textbf{Implementation Details.}
    We describe the detailed implementation of SVGFusion, including data standardization, hyperparameter settings, training configurations, and an inference efficiency analysis.

    \item \cref{sec:supp_svg_data}: \textbf{SVGX-Dataset Details.}
    We provide a comprehensive overview of our newly introduced SVGX-Dataset, covering data representation protocols, preprocessing pipelines, and cleaning procedures used to ensure data quality.

    \item \cref{sec:supp_vs-dit}: \textbf{VS-DiT Architecture.}
    We present the architectural specifications of the Vector Space Diffusion Transformer (VS-DiT), explaining its core components and the rationale behind our design choices.

    \item \cref{sec:supp_compare_llm}: \textbf{Comparison with Autoregressive Language Models.}
    We analyze the critical differences between our approach and language model-based methods, highlighting how SVGFusion leverages a \textbf{holistic diffusion paradigm} to mitigate the error accumulation and structural inconsistency inherent in autoregressive generation.

    \item \cref{sec:supp_primitive_type}: \textbf{Expanded SVG Primitive Support.}
    We demonstrate that SVGFusion supports a broader range of SVG primitives than existing methods and detail our \textbf{canonicalization strategy} for unifying heterogeneous geometric commands.

    \item \cref{sec:supp_results}: \textbf{Additional Qualitative Results.}
    We provide extensive visualizations of the diffusion process and generated samples, showcasing SVGFusion’s capabilities in high-fidelity icon synthesis, \textbf{direct attribute editing}, and \textbf{vector recomposition}.
\end{itemize}

\section{Implementation Details}
\label{sec:supp_implement_detail}

\noindent\textbf{Data Standardization and Optimization.} 
To ensure consistency across all SVG data, we adopted relative positional coordinates. 
The model parameters were initialized randomly and optimized using the AdamW optimizer (with $\beta_1=0.9$, $\beta_2=0.999$) at an initial learning rate of $5 \times 10^{-5}$. 
The learning rate was warmed up over the first 2,000 steps and then decayed to $1 \times 10^{-6}$ following a cosine schedule. 
Additionally, we applied a weight decay of $0.1$ for regularization and constrained gradients by clipping their norms to a maximum value of $1.0$.
Furthermore, input SVG embeddings were normalized into the $[-1, 1]$ range to stabilize the training process.

\noindent\textbf{Architecture and Positional Embeddings.}
We utilized the Transformer~\cite{transfromer_ashish_2017} architecture as the fundamental building block for VP-VAE. 
Both the encoders and decoders consist of 4 layers with a hidden dimension of 512.
To further enhance the model’s ability to capture sequential dependencies, we integrated Rotary Position Embeddings (RoPE)~\cite{roformer_su_2024}, a technique widely used in advanced large language models (LLMs)~\cite{ChatGPT,llama_Touvron_2023}. 
Although SVGs represent 2D visual content, we treat the SVG tensor as a 1D sequence of primitives.
RoPE effectively encodes the positional relationships within this sequence, allowing the model to better understand the temporal logic and structural progression (e.g., drawing order) of SVG creation.

\noindent\textbf{Training Configurations.}
To investigate scaling trends, we trained VS-DiT models at three different sizes: 0.16B, 0.37B, and 0.76B parameters.
We trained the VP-VAE for 1,000k steps using a total batch size of 512 across 8 NVIDIA A800 GPUs, which required approximately two days. 
Subsequently, leveraging the frozen VP-VAE, we trained the VS-DiT for 500k steps with a batch size of 512 on 8 H800 GPUs, taking approximately three days.

\noindent\textbf{Inference Efficiency.}
In the sampling phase, we utilized DDIM~\cite{ddim_song_2021} for 150-step denoising by default. 
For efficient sampling, we also support the DPM-Solver~\cite{dpmsolver_lu_2022} with 20 steps. 
As shown in Table~1 of the main paper, our method demonstrates superior inference speed compared to optimization-based methods. specifically, with the DPM-Solver, the inference time is reduced to \textit{24s (Small), 28s (Base), and 36s (Large)}, significantly outperforming optimization-based baselines which typically require several minutes (e.g., 35-47 minutes).

\begin{figure*}
\centering
\includegraphics[width=1.0\textwidth]{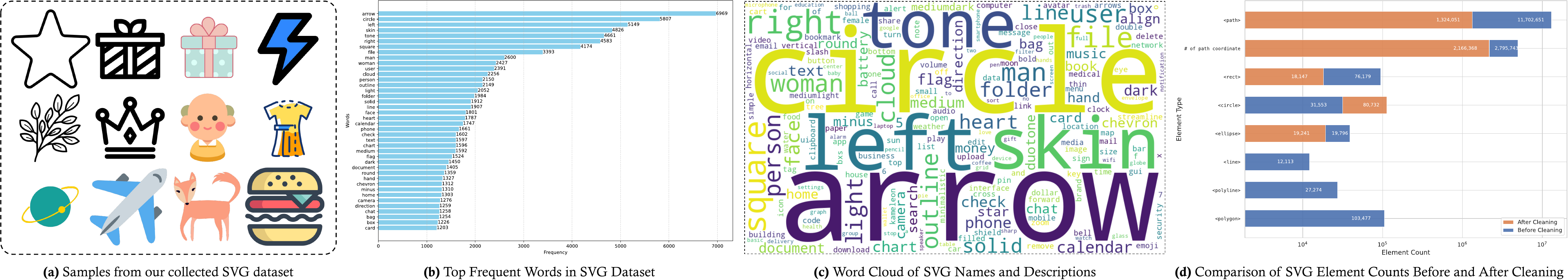}
\vspace{-1.5em}
\caption{
\textbf{Overview and analysis of the proposed SVGX-Dataset.} 
\textbf{(a)} Sample visualizations showing the diversity of styles, ranging from outlined icons to filled flat illustrations.
\textbf{(b)} The top-40 most frequent words in dataset entry names, indicating a balanced distribution of common objects and symbols.
\textbf{(c)} Word cloud of names and descriptions, highlighting key semantic themes such as geometry (``circle'', ``square'') and UI elements (``file'', ``arrow'').
\textbf{(d)} Quantitative comparison of SVG element counts before and after our cleaning pipeline. Note the significant reduction in \texttt{<path>} coordinates and redundant tags (displayed on a log scale), demonstrating the efficiency of our preprocessing in removing noise while retaining structural information.
}
\label{fig:dataset_analysis}
\vspace{-0.5em}
\end{figure*}
\begin{figure*}[t]
\centering
\includegraphics[width=1.0\linewidth]{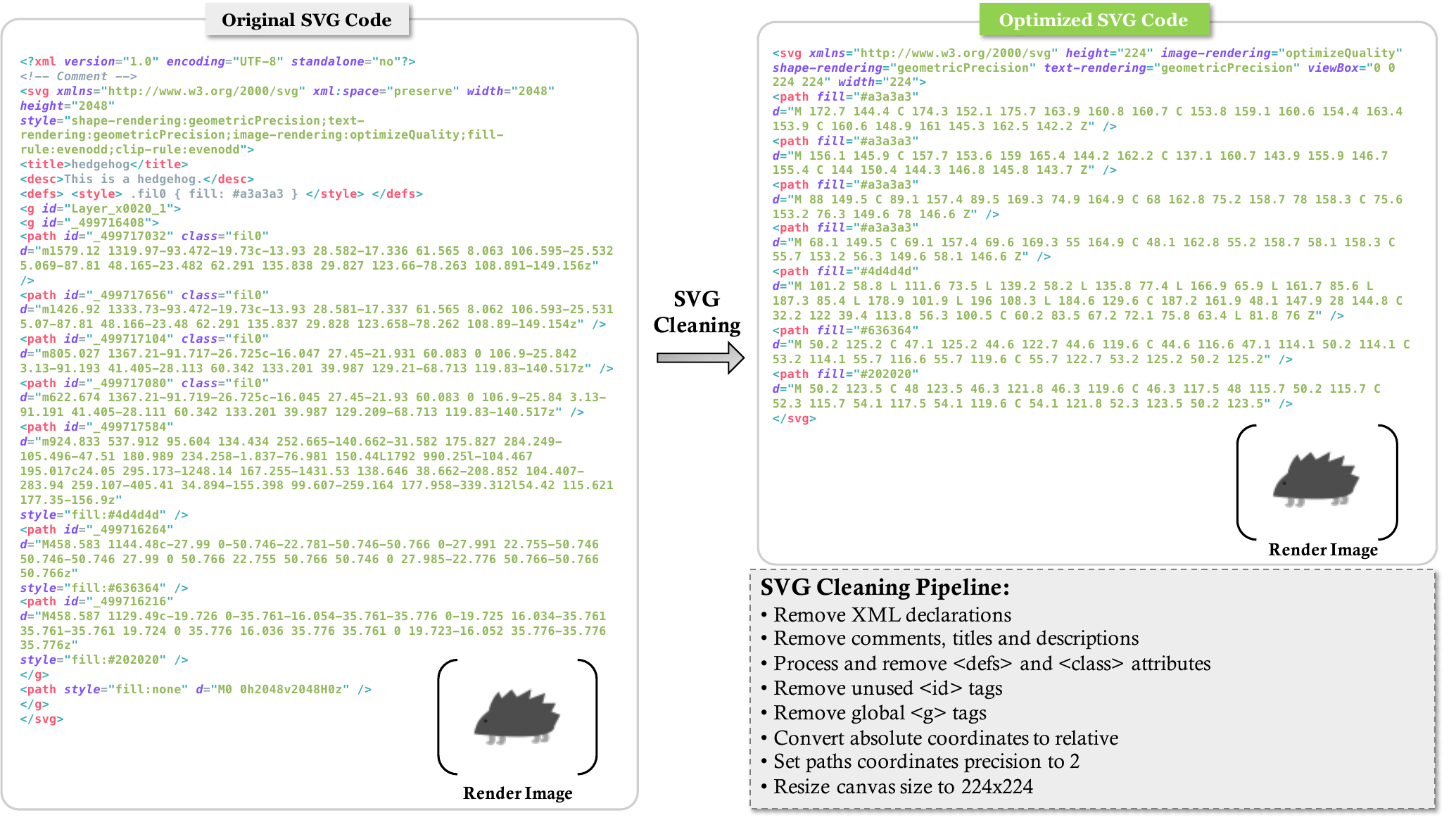}
\vspace{-2em}
\caption{
\textbf{Overview of the SVG cleaning pipeline.} 
We transform raw, verbose SVG files (Left) into a compact, optimized representation (Right) suitable for model training. 
The pipeline systematically removes non-structural noise, including XML headers, comments, and unused metadata. 
Crucially, it flattens the structural hierarchy by eliminating \texttt{<defs>} and \texttt{<class>} attributes and normalizes geometry by converting absolute coordinates to relative ones with rounded precision. 
The canvas resolution is standardized to $224\times224$. 
As shown, the optimized code is significantly more concise and token-efficient, yet it preserves the exact visual fidelity of the original image (see Render Image).
}
\vspace{-0.5em}
\label{fig:supp_svg_clean}
\end{figure*}

\section{SVGX-Dataset: Curation, Preprocessing, and Analysis}
\label{sec:supp_svg_data}

\noindent\textbf{Dataset Curation and Sources.}\quad
We constructed the \textit{SVGX-Dataset}, a large-scale corpus designed for high-quality vector graphic generation. 
The dataset comprises approximately \textbf{240k} samples, focusing on \textit{emoji} and \textit{icon} styles to ensure structural clarity and aesthetic appeal.
Data sources include Twemoji-Color-Font~\cite{twitter_emoji}, Noto-Emoji~\cite{notoemoji_google_2022}, FluentUI-Emoji~\cite{fluent_ms}, SVG-Repo~\cite{svgrepo}, and Reshot~\cite{reshot_data}.
Specifically, the three emoji datasets contribute approximately 4,000 samples each. 
Reshot provides an additional 30,000 high-quality icons, while the majority—approximately 200,000 SVGs—are sourced from SVGRepo.
This diverse sourcing strategy ensures a rich distribution of semantic categories and artistic styles.

\noindent \textbf{Data Representation and Diversity.}\quad
Our dataset spans a broad spectrum of visual complexity, ranging from simple geometric shapes to intricate illustrations.
Structurally, the dataset incorporates a mix of SVG primitives: it utilizes B\'ezier curves (\texttt{<path>}) for complex contours and basic shapes (\texttt{<circle>}, \texttt{<rect>}, \texttt{<ellipse>}) for regular geometric elements.
As visualized in Fig.~\ref{fig:dataset_analysis}(a), the collection covers diverse themes—including nature, objects, symbols, animals, and food—rendered in both monochromatic and vibrant color palettes. 
This semantic and structural diversity allows the model to learn robust representations applicable to various design scenarios.

\noindent\textbf{Automated Cleaning Pipeline.}\quad
Raw SVG files crawled from the web often contain significant noise, such as editor-specific metadata, redundant definitions, and invisible elements. Direct training on such noisy data leads to inefficient token utilization and hinders model convergence.
To address this, we developed a lossless preprocessing pipeline.
As shown in the comparison in Fig.~\ref{fig:dataset_analysis}(d), raw SVGs often contain millions of redundant path coordinates and unused definitions.
Our pipeline performs the following optimizations: 
(1) removing XML declarations, comments, and metadata; 
(2) stripping unused \texttt{<defs>} and invisible groups; 
(3) converting absolute coordinates to relative ones to enhance translation invariance; 
(4) rounding coordinate precision to two decimal places; 
and (5) standardizing the canvas size to $224\times224$.
This process significantly reduces the sequence length while preserving visual fidelity.

\noindent\textbf{Statistical Analysis.}\quad
We analyze the linguistic and structural characteristics of the dataset in Fig.~\ref{fig:dataset_analysis}.
The frequency analysis of entry names (Fig.~\ref{fig:dataset_analysis}(b)) and the semantic word cloud (Fig.~\ref{fig:dataset_analysis}(c)) highlight a predominance of geometric (``circle'', ``square'') and directional (``arrow'', ``left'', ``right'') concepts, confirming the dataset's suitability for iconographic generation tasks.
Crucially, Fig.~\ref{fig:dataset_analysis}(d) demonstrates the efficacy of our cleaning pipeline, showing a dramatic reduction in the total count of SVG elements (e.g., \texttt{<path>} coordinates dropped from $\sim$11.7M to $\sim$1.8M), proving that our preprocessing yields a highly compact and information-dense representation for model training.

\begin{figure}[t]
\centering
\includegraphics[width=0.8\linewidth]{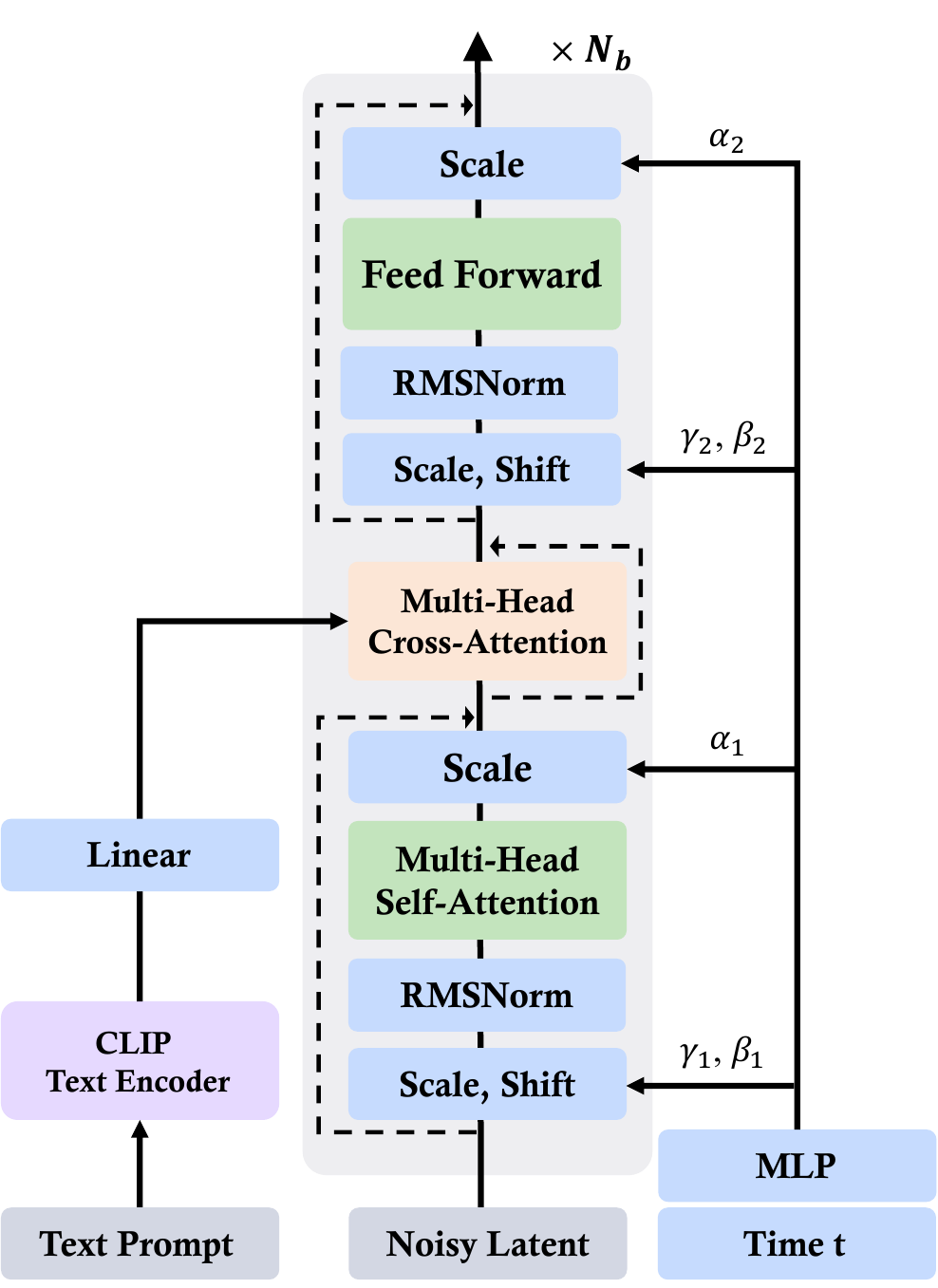}
\caption{
\textbf{The architecture of the VS-DiT Block.} 
The architecture of the VS-DiT Block, where a multi-head cross-attention module is incorporated into each block to inject textual conditions. The text prompt is first processed through a CLIP text encoder and a linear transformation before being integrated to the model. The backbone follows a pre-norm topology using RMSNorm for feature normalization. Time-dependent conditioning is introduced through an MLP that predicts adaptive parameters ($\gamma, \beta$) for feature scaling and shifting, as well as gating scalars ($\alpha$) to regulate the residual contributions of the multi-head self-attention and feed-forward layers. The block is repeated $N_b$ times to refine the latent representation.
} \label{fig:supp_arch}
\vspace{-0.5em}
\end{figure}

\section{Architecture Details of the VS-DiT Block}
\label{sec:supp_vs-dit}

\noindent\textbf{Block Architecture.}\quad 
The core component of our SVGFusion is the Vector Space Diffusion Transformer (VS-DiT) Block. 
As illustrated in Fig.~\ref{fig:supp_arch}, we build upon the standard DiT architecture~\cite{dit_peebles_2023} by incorporating a \textbf{Multi-Head Cross-Attention} module. 
This module is strategically positioned between the self-attention layer and the feed-forward network to facilitate robust interaction with textual conditions. 
Specifically, text prompts are first encoded by a frozen CLIP encoder and projected via a linear layer before being injected into the diffusion process.
To handle temporal dependencies, we employ \textit{adaptive layer normalization} (AdaLN). 
A time-dependent MLP predicts the scale and shift parameters ($\gamma, \beta$) based on the diffusion timestep $t$, modulating the normalized features. 
Furthermore, to stabilize training, we introduce learnable gating scalars ($\alpha_1, \alpha_2$), initialized to zero, which regulate the residual contributions of the attention and feed-forward blocks.

\noindent\textbf{Model Scalability and Configurations.}\quad
To investigate the scaling laws in vector graphic generation, we implement a stack of $N_b$ VS-DiT blocks, operating at a hidden dimension size $d$. 
Following the design philosophy of DiT, we define three model variants—\textit{VS-DiT-S} (Small), \textit{VS-DiT-B} (Base), and \textit{VS-DiT-L} (Large)—by jointly scaling the depth $N_b$, hidden dimension $d$, and the number of attention heads. 
These configurations span a computational spectrum from 1.4 to 19.9 GFLOPs, allowing for a comprehensive analysis of performance versus computational cost. 
Detailed hyperparameter specifications for each variant are provided in Table~\ref{tab:model_size}.

\begin{table}[t]
\centering
\resizebox{0.8\linewidth}{!}{
\begin{tabular}{ccccc}
\toprule
Model&Layer $N_b$&Hidden size $d$&Heads&Gflops\\
\midrule
VS-DiT S &12&384&6&1.4\\
\midrule
VS-DiT B &12&768&12&5.6 \\
\midrule
VS-DiT L &24&1024&16&19.9 \\
\bottomrule
\end{tabular}
}
\vspace{-0.5em}
\caption{
\textbf{Hyperparameter specifications for VS-DiT variants.} 
We follow standard Transformer scaling configurations to evaluate model performance across different capacities.
} \label{tab:model_size}
\vspace{-0.5em}
\end{table}

\begin{table*}
\centering
\resizebox{1.0\linewidth}{!}{
\begin{tabular}{p{0.3\textwidth} p{0.2\textwidth} p{0.5\textwidth} c}
\toprule
\makecell[tl]{\textbf{SVG Primitives} \\ \textbf{(Element/Command)}} 
& \textbf{Argument} & \textbf{Explanation} & \textbf{Example} \\
\midrule
\textbf{\texttt{<circle>}} & $r, cx, cy$ & The \texttt{<circle>} element is used to create a circle with center at $(cx, cy)$ and radius $r$. & 
\begin{tikzpicture}[scale=0.5, baseline=(current bounding box.center)]
    \draw[thick] (2,2) circle (1.5cm);
    \filldraw[cyan] (2,2) circle (0.05cm) node[below right, font=\tiny] {$(cx, cy)$};
    \draw[dashed] (2,2) -- (3.5,2) node[midway, above, font=\tiny] {$r$};
\end{tikzpicture} \\
\midrule
\textbf{\texttt{<ellipse>}} & $rx, ry, cx, cy$ & The \texttt{<ellipse>} element is used to create an ellipse with center at $(cx, cy)$, and radii $rx$ and $ry$. & 
\begin{tikzpicture}[scale=0.5, baseline=(current bounding box.center)]
    \draw[thick] (2,2) ellipse (2cm and 1cm);
    \filldraw[cyan] (2,2) circle (0.05cm) node[below right, font=\tiny] {$(cx, cy)$};
    \draw[dashed] (2,2) -- (4,2) node[midway, above, font=\tiny] {$rx$};
    \draw[dashed] (2,2) -- (2,3) node[midway, right, font=\tiny] {$ry$};
\end{tikzpicture} \\
\midrule
\textbf{\texttt{<rect>}} & $rx, ry, cx, cy$ & The \texttt{<rect>} element is used to create a rectangle, optionally with rounded corners if $rx$ and $ry$ are specified. The center is at $(cx, cy)$. & 
\begin{tikzpicture}[scale=0.5, baseline=(current bounding box.center)]
    \draw[thick, rounded corners=0.2cm] (1,1) rectangle (3,3);
    \filldraw[cyan] (2,2) circle (0.05cm) node[below right, font=\tiny] {$(cx, cy)$};
    \draw[dashed] (1,2) -- (3,2) node[midway, above, font=\tiny] {width};
    \draw[dashed] (2,1) -- (2,3) node[midway, right, font=\tiny] {height};
\end{tikzpicture} \\
\midrule
\midrule
\texttt{<path>} Move To (\texttt{M}) & $\mu_3, \nu_3$ & Moves the cursor to the specified point $(\mu_3, \nu_3)$. & 
\raisebox{-0.5\height}{\includegraphics[width=0.15\textwidth]{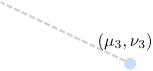}} \\
\midrule
\texttt{<path>} Line To (\texttt{L}) & $\mu_3, \nu_3$ & Draws a line segment from the current point to $(\mu_3, \nu_3)$. & \raisebox{-0.5\height}{\includegraphics[width=0.15\textwidth]{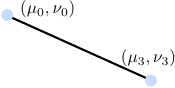}} \\
\midrule
\texttt{<path>} Cubic Bézier (\texttt{C}) & $\mu_1, \nu_1, \mu_2, \nu_2, \mu_3, \nu_3$ & Draws a cubic Bézier curve with control points $(\mu_1, \nu_1)$, $(\mu_2, \nu_2)$, and endpoint $(\mu_3, \nu_3)$. & \raisebox{-0.5\height}{\includegraphics[width=0.15\textwidth]{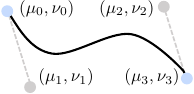}} \\
\midrule
\texttt{<path>} \textbf{Quadratic Bézier (\texttt{Q})}
& $\mu_1, \nu_1, \mu_3, \nu_3$ & Draws a quadratic Bézier curve with control points $(\mu_1, \nu_1)$ and endpoint $(\mu_3, \nu_3)$. & \raisebox{-0.5\height}{\includegraphics[width=0.15\textwidth]{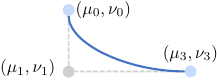}} \\
\midrule
\texttt{<path>} \textbf{Elliptical Arc (\texttt{A})}
& $\texttt{rx}, \texttt{ry}, \texttt{rotate}, \newline \texttt{LargeArcFlag}, \newline \texttt{SweepFlag}, \mu_3, \nu_3$ 
& Draws an elliptical arc from the current point to $(\mu_3, \nu_3)$. The ellipse has radii \texttt{rx}, \texttt{ry}, rotated by \texttt{rotate} degrees. \texttt{LargeArcFlag} and \texttt{SweepFlag} control the arc direction. & \raisebox{-0.5\height}{\includegraphics[width=0.15\textwidth]{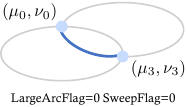}} \\
\midrule
\texttt{<path>} Close Path (\texttt{Z}) & $\varnothing$ & Closes the path by moving the cursor back to the path’s starting position $(\mu_0, \nu_0)$. & \raisebox{-0.5\height}{\includegraphics[width=0.15\textwidth]{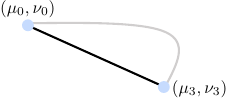}} \\
\midrule
\midrule
\texttt{<SOS>} & $\varnothing$ & Special token indicating the start of an SVG sequence. & N/A \\
\midrule
\texttt{<EOS>} & $\varnothing$ & Special token indicating the end of an SVG sequence. & N/A \\
\bottomrule
\end{tabular}
}
\vspace{-0.5em}
\caption{
\textbf{Detailed specification of supported SVG primitives.} 
Unlike prior methods~\cite{deepsvg_carlier_2020, iconshop_wu_2023} that typically restrict representation to basic paths (Commands \texttt{M, L, C}), SVGFusion natively supports a significantly expanded vocabulary. 
The table is divided into \textbf{Element-level primitives} (top three rows) and \textbf{Path-level commands} (bottom rows). 
Items highlighted in \textbf{bold} (e.g., \texttt{<circle>}, \texttt{<rect>}, \texttt{Q}, \texttt{A}) represent our specific extensions over standard baselines. 
This expanded support allows the model to preserve specific geometric structures (like perfect circles or arcs) without approximating them as generic B\'ezier curves.
} \label{tab:svg_commands}
\vspace{-0.5em}
\end{table*}

\section{Comparison with Autoregressive Language Models}
\label{sec:supp_compare_llm}

We contrast SVGFusion with language model-based methods (e.g., DeepSVG~\cite{deepsvg_carlier_2020}, IconShop~\cite{iconshop_wu_2023}) from two critical perspectives: generation paradigm and representation efficiency.

\noindent\textbf{Holistic Generation vs. Sequential Error Propagation.}\quad
Language model-based approaches treat SVG generation as a standard next-token prediction task. 
However, this autoregressive (AR) nature inherently limits their performance in vector graphics. 
As visually demonstrated in the top row of Fig.~\ref{fig:supp_llm_vs_ddpm}, the prediction of each coordinate or command is strictly conditioned on the accuracy of preceding tokens. 
This leads to \textit{irreversible error accumulation}: a single misplaced coordinate or malformed syntax tag early in the sequence can trigger a cascading failure, resulting in unclosed paths, disjointed shapes, or spatial misalignment. The model lacks a "global view" to correct earlier mistakes during generation.

\noindent\textbf{Iterative Refinement vs. One-Pass Generation.}\quad
In contrast, SVGFusion employs a diffusion-based paradigm that synthesizes the entire graphic in parallel. 
As shown in the bottom row of Fig.~\ref{fig:supp_llm_vs_ddpm}, our process begins with a global Gaussian noise distribution. The model iteratively denoises all primitives simultaneously, refining the global structure and local details essentially at the same time. 
This allows SVGFusion to maintain \textit{global visual coherence}, as the model can adjust the relationship between shapes dynamically throughout the denoising steps, effectively mitigating the "drift" issues common in AR methods.

\noindent\textbf{Representation Efficiency.}\quad
Finally, LLM-based methods typically process SVG as raw text (XML), which is highly verbose and token-inefficient. A simple shape may require hundreds of tokens to describe, diluting the semantic density. 
Our method, conversely, operates in a compact, structured latent space. By embedding primitives into dense vectors, SVGFusion achieves higher information density and computational efficiency, avoiding the burden of processing lengthy, redundant XML syntax.

\begin{figure}[th]
\centering
\includegraphics[width=1\linewidth]{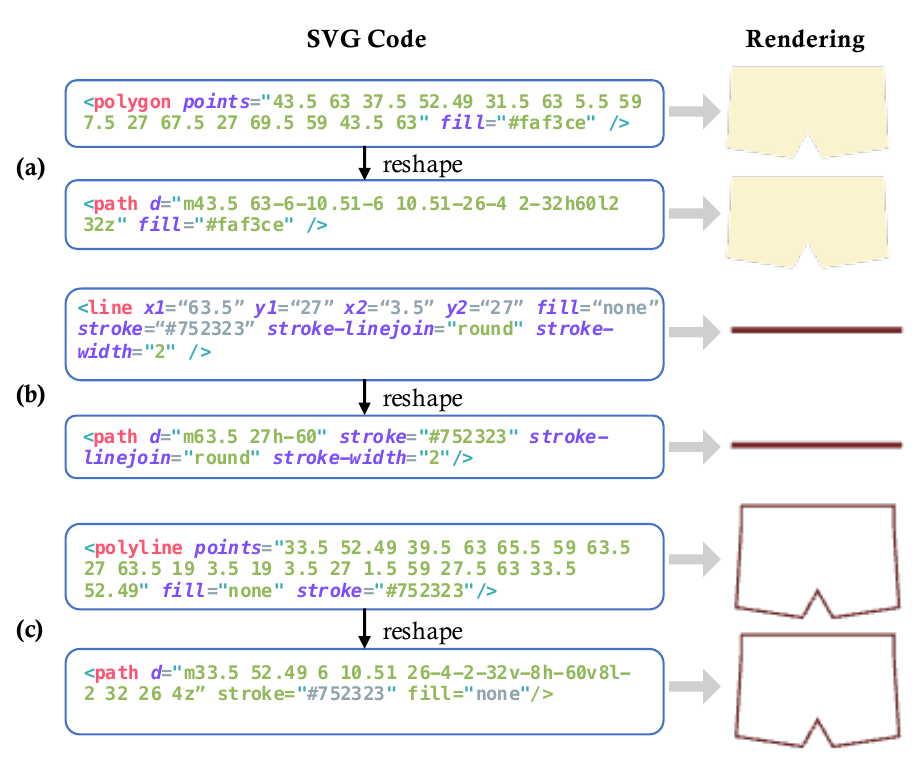}
\vspace{-2em}
\caption{
\textbf{Unified reshaping of SVG primitives into path format.}
This figure illustrates how heterogeneous SVG primitives—(a) \texttt{<polygon>}, (b) \texttt{<line>}, and (c) \texttt{<polyline>}—are converted into an equivalent \texttt{<path>} representation without any loss of geometric fidelity. 
Our reshaping procedure unifies structurally different primitives into a single, canonical path format, enabling consistent tokenization, representation learning, and downstream reasoning within our model. By eliminating format-level discrepancies while preserving visual rendering exactly, this normalization significantly simplifies SVG parsing and allows the model to operate on a more compact and expressive design space.
} \label{fig:supp_svg_reshape}
\vspace{-0.5em}
\end{figure}
\section{Expanded SVG Primitive Support and Canonicalization}
\label{sec:supp_primitive_type}

Standard SVG syntax encompasses a rich set of primitives and complex compositional rules. 
However, prior language model-based approaches~\cite{deepsvg_carlier_2020, deepvecfont_wang_2021, deepvecfontv2_wang_2023, iconshop_wu_2023, strokenuwa_tang_2024} often resort to extreme simplification, limiting the representation to a single element-level primitive (\texttt{<path>}) and a subset of commands (Move \texttt{M}, Line \texttt{L}, Cubic B\'ezier \texttt{C}). 
Such reductionist approaches fail to capture the semantic intent of human designers and the structural diversity of real-world vector graphics.
To bridge this gap, SVGFusion introduces a more comprehensive primitive handling strategy, consisting of two key components: expanded native support and unified canonicalization.

\noindent\textbf{Expanded Native Support.}\quad
As detailed in Table~\ref{tab:svg_commands}, our model natively encodes a broader spectrum of geometric primitives. 
Beyond generic paths, we explicitly support semantically distinct shapes including \texttt{<circle>}, \texttt{<rect>}, and \texttt{<ellipse>}. 
Within the path data structure, we extend support to advanced commands such as Quadratic B\'ezier curves (\texttt{Q}), Elliptical Arcs (\texttt{A}), and Path Closure (\texttt{Z}). 
This allows the model to learn and generate specific geometric structures (e.g., perfect circles or rounded rectangles) using their most efficient and semantically correct representations, rather than approximating them with generic B\'ezier paths.

\noindent\textbf{Unified Canonicalization Strategy.}\quad
To handle the heterogeneity of SVG syntax without exploding the vocabulary size, we implement a lossless reshaping pipeline (illustrated in Fig.~\ref{fig:supp_svg_reshape}) to canonicalize redundant primitives.
\begin{itemize}
\item \textbf{Element Normalization:} Primitives such as \texttt{<line>}, \texttt{<polygon>}, and \texttt{<polyline>} are mathematically converted into their equivalent \texttt{<path>} representations. This ensures visual fidelity while unifying the input format.
\item \textbf{Command Normalization:} Shorthand path commands are expanded to their explicit forms to reduce ambiguity. Specifically, Horizontal (\texttt{H}) and Vertical (\texttt{V}) lines are mapped to Line To (\texttt{L}); Smooth Cubic (\texttt{S}) and Smooth Quadratic (\texttt{T}) B\'eziers are converted to explicit Cubic (\texttt{C}) and Quadratic (\texttt{Q}) curves, respectively.
\end{itemize}
This strategy streamlines the token space and ensures consistency during training without sacrificing geometric precision.


\begin{figure}
\centering
\includegraphics[width=1\linewidth]{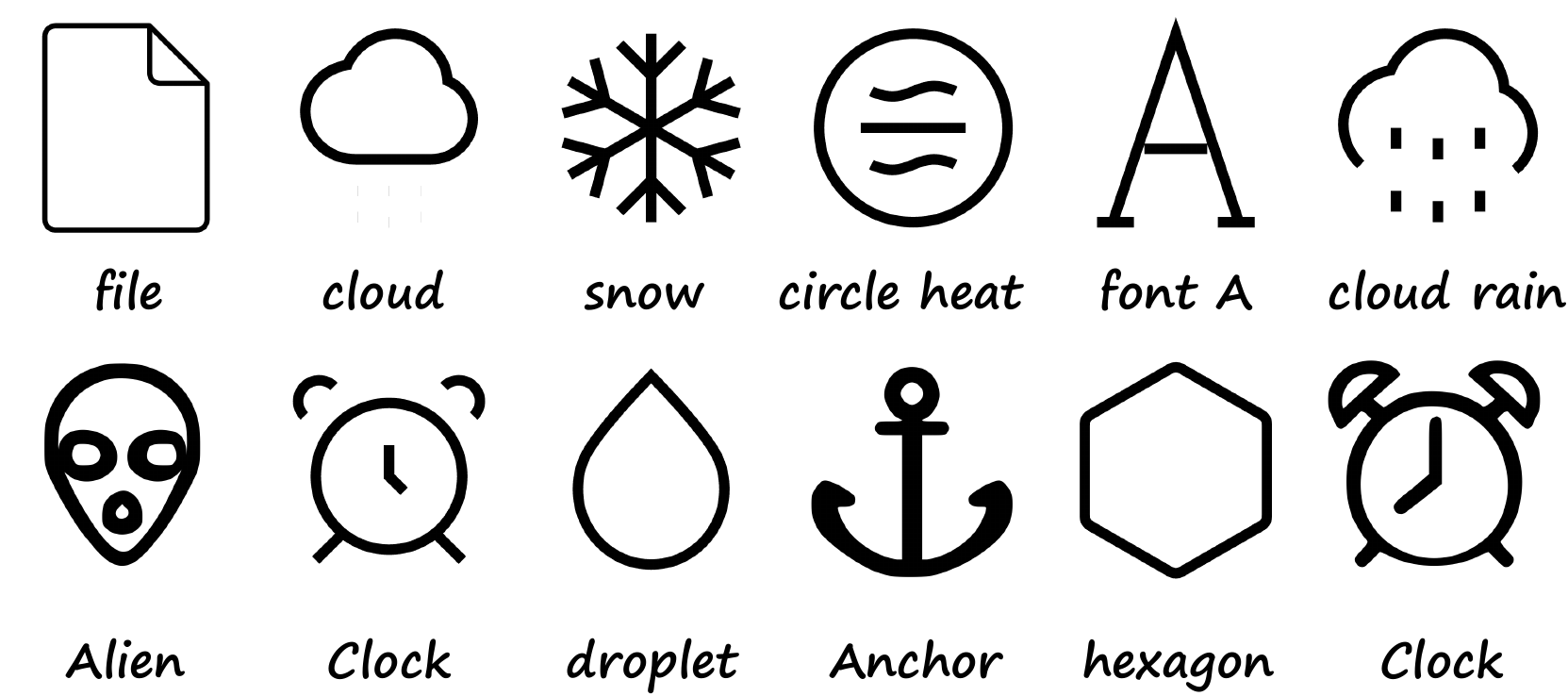}
\vspace{-1.5em}
\caption{\textbf{Gallery of black-and-white icons generated by SVGFusion.} 
The model effectively synthesizes topologically clean and semantically recognizable symbols across various categories (weather, time, UI elements). Note the clean strokes and closed shapes, which are ideal for direct use in vector design applications.}
\label{fig:supp_icon}
\vspace{-0.5em}
\end{figure}
\begin{figure}[th]
\centering
\includegraphics[width=1\linewidth]{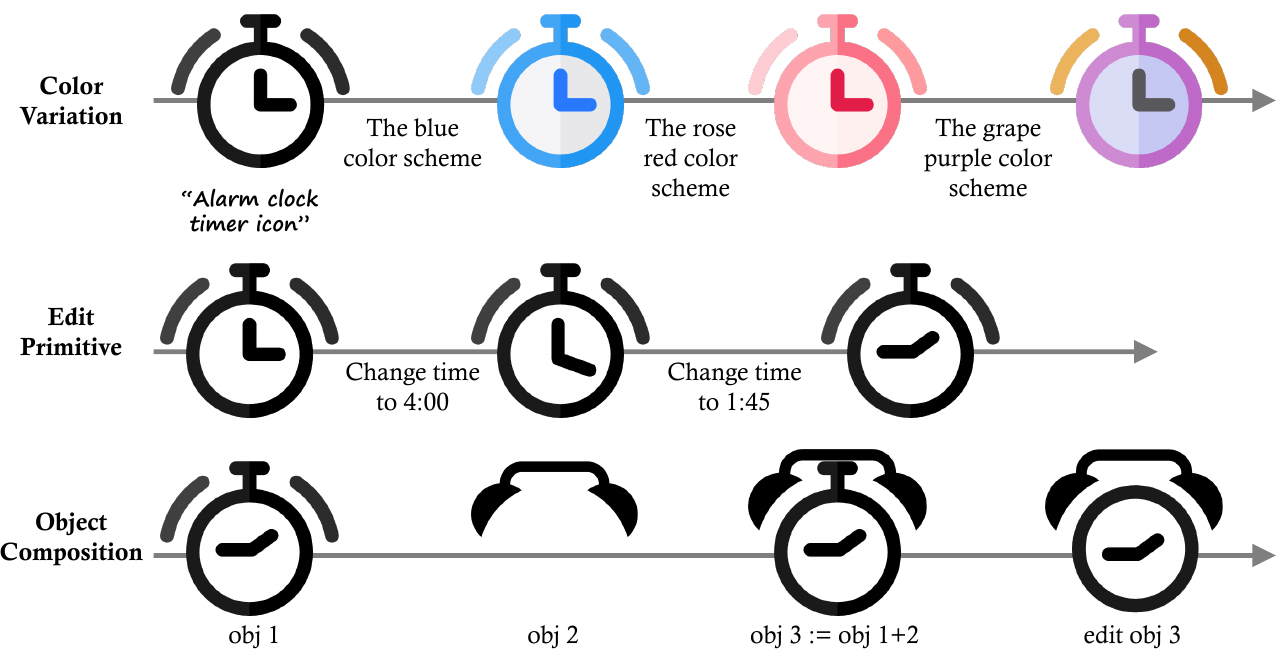}
\vspace{-1.5em}
\caption{\textbf{Demonstration of SVG editability and recomposition.} 
The structural integrity of SVGFusion's outputs allows for precise downstream editing.
\textit{Top (Color Variation):} Users can modify the color scheme simply by changing the \texttt{fill} attributes in the code, without needing to re-generate the image.
\textit{Middle (Edit Primitive):} Geometric parameters (e.g., clock hands) can be adjusted via coordinate modification.
\textit{Bottom (Vector Recomposition):} The generated SVGs are modular, allowing distinct components to be decomposed and recombined (e.g., merging the clock body from Obj 1 with hands from Obj 2) to form new vector graphics.}
\label{fig:supp_edit}
\vspace{-0.5em}
\end{figure}

\section{Additional Qualitative Results}
\label{sec:supp_results}

\noindent\textbf{High-Fidelity Icon Generation.}\quad
As illustrated in Fig.~\ref{fig:supp_icon}, SVGFusion excels at synthesizing clean, black-and-white icon-style SVGs. 
Unlike raster-based methods that may produce jagged edges or blurry artifacts, our model generates crisp, resolution-independent geometries. 
These icons are typically composed of a minimal number of primitives, demonstrating the model's ability to capture the essence of a subject with expressive yet concise design characteristics.

\noindent\textbf{Direct Attribute Editability.}\quad
Figure~\ref{fig:supp_edit} highlights the native editability of SVGs produced by our framework. 
Because SVGFusion generates structured XML code rather than pixel grids, the outputs possess a clear hierarchical organization. 
This allows users to perform \textbf{direct code-level manipulation}: primitive properties such as fill color, stroke width, and geometric coordinates can be modified precisely without requiring complex image inpainting or regeneration. 
This seamless customization capability is a significant advantage over pixel-based diffusion models.

\noindent\textbf{Vector Recomposition.}\quad
Beyond attribute modification, the disentangled nature of our generated primitives supports complex vector recomposition.
As shown in the bottom row of Fig.~\ref{fig:supp_edit}, users can efficiently repurpose synthesized elements—extracting specific components (e.g., the body of a clock) and recombining them with other objects to construct entirely novel designs. 
This modularity not only showcases the flexibility of our framework but also significantly enhances the reusability of the generated assets in real-world design workflows.

{
    \small
    \bibliographystyle{ieeenat_fullname}
    \bibliography{main}
}

\end{document}